\documentclass[nohyperref]{article}

\usepackage[accepted]{icml2022}

\usepackage{amsmath}
\usepackage{amsmath,amsfonts,bm}

\def\eqref#1{(\ref{#1})}
\def\1{\bm{1}}

\DeclareMathAlphabet{\mathsfit}{\encodingdefault}{\sfdefault}{m}{sl}
\SetMathAlphabet{\mathsfit}{bold}{\encodingdefault}{\sfdefault}{bx}{n}

\DeclareMathOperator*{\argmin}{arg\,min}

\usepackage[utf8]{inputenc} % allow utf-8 input
\usepackage[T1]{fontenc}    % use 8-bit T1 fonts
\usepackage{hyperref}       % hyperlinks
\usepackage{url}            % simple URL typesetting
\usepackage{booktabs}       % professional-quality tables
\usepackage{amsfonts}       % blackboard math symbols
\usepackage{nicefrac}       % compact symbols for 1/2, etc.
\usepackage{microtype}      % microtypography

 \usepackage{graphicx}
\usepackage{subfigure} 
\usepackage{pifont}
\usepackage{adjustbox}
\usepackage{lipsum}
\usepackage{color}
\usepackage{wrapfig}
\usepackage{booktabs}
\usepackage[textsize=scriptsize]{todonotes}
\usepackage{multirow,mathtools} 
\usepackage{threeparttable}
\usepackage{xcolor}

\definecolor{lightblue}{rgb}{0.68, 0.85, 0.9}
\definecolor{lightgreen}{rgb}{0.56, 0.93, 0.56}
\definecolor{lightskyblue}{rgb}{0.53, 0.81, 0.98}
\definecolor{non-photoblue}{rgb}{0.64, 0.87, 0.93}
\definecolor{magicmint}{rgb}{0.67, 0.94, 0.82}
\definecolor{mossgreen}{rgb}{0.68, 0.87, 0.68}
\definecolor{salmon}{rgb}{1.0, 0.55, 0.41}
\definecolor{babypink}{rgb}{0.96, 0.76, 0.76}

\newtheorem{myprop}{\bf{Proposition}}
\newtheorem{mycor}{\bf{Corollary}}

\DeclareMathOperator*{\minimize}{\text{minimize}}
\DeclareMathOperator*{\maximize}{\text{maximize}}
\DeclareMathOperator*{\st}{\text{subject to}}
\DeclareMathAlphabet\mathbfcal{OMS}{cmsy}{b}{n}
\newcommand{\Def}[0]{\mathrel{\mathop:}=}

\def\remark{\addtocounter{remark}{1}\def\@currentlabel{\theremark}%
\emph{Remark~\theremark}. } \makeatother

\newcounter{remark}

\def\btheta{\boldsymbol{\theta}}
\def\bdelta{\boldsymbol{\delta}}

\def\bdelta{\boldsymbol{\delta}}

\usepackage{color, colortbl}
\definecolor{Gray}{gray}{0.93}
\definecolor{Orange}{rgb}{1,0.5,0}
\definecolor{DGray}{gray}{0.83}
\definecolor{LightCyan}{rgb}{0.88,1,1}

\usepackage[most]{tcolorbox}
\newtcolorbox{mybox}[2][]{%
  attach boxed title to top center
               = {yshift=-8pt},
  colback      = Gray,
  colframe     = black,
  fonttitle    = \bfseries,
  colbacktitle = white,
  title        = #2,#1,
  enhanced,
}

\DeclarePairedDelimiterX{\inp}[2]{\langle}{\rangle}{#1, #2}

\makeatletter
\newcommand*{\rom}[1]{\expandafter\@slowromancap\romannumeral #1@}
\makeatother
\newcommand{\mycomment}[1]{}

\usepackage{multirow}

\newcommand{\IG}{{\text{IG}}}
\newcommand{\FBAT}{{\textsc{Fast-BAT}}}
\newcommand{\FAT}{{\textsc{Fast-AT}}}
\newcommand{\FATGA}{{\textsc{Fast-AT-GA}}}
\newcommand{\AT}{{\textsc{AT}}}
\newcommand{\ATF}{{\textsc{PGD-2-AT}}}

\newcommand{\bs}{\textsc{BackSmooth}}
\newcommand{\free}{\textsc{Free-AT}}
\newcommand{\ATTA}{\textsc{ATTA}}
\newcommand{\YOPO}{\textsc{YOPO}}
\newcommand{\PGDS}{\textsc{PGD-7-AT}}

\newcommand{\RV}[1]{\textcolor{black}{#1}}

\definecolor{light-gray}{gray}{0.8}
\newcommand{\mathHigh}[1]{\colorbox{Gray}{$\displaystyle #1$}}
\newcommand{\mycolorrow}{\cellcolor{Gray}}

\icmltitlerunning{Revisiting and Advancing Fast Adversarial Training Through the Lens of Bi-Level Optimization}

\begin{document}

\twocolumn[
\icmltitle{Revisiting and Advancing Fast Adversarial Training \\ Through the Lens of Bi-Level Optimization}



\icmlsetsymbol{equal}{*}

\begin{icmlauthorlist}
\icmlauthor{Yihua Zhang}{msu,equal}
\icmlauthor{Guanhua Zhang}{ucsb,equal}
\icmlauthor{Prashant Khanduri}{umn}
\icmlauthor{Mingyi Hong}{umn}
\icmlauthor{Shiyu Chang}{ucsb}
\icmlauthor{Sijia Liu}{msu,ibm}
\end{icmlauthorlist}

\icmlaffiliation{msu}{
% Computer Science \& Engineering, 
Michigan State University}
\icmlaffiliation{ucsb}{
% Computer Science, 
UC Santa Barbara}
\icmlaffiliation{umn}{
% Electrical \& Computer Engineering,
University of Minnesota}
\icmlaffiliation{ibm}{MIT-IBM Watson AI Lab, IBM Research}

\icmlcorrespondingauthor{Yihua Zhang}{zhan1908@msu.edu}
\icmlcorrespondingauthor{Guanhua  Zhang}{guanhua@ucsb.edu}

\icmlkeywords{Machine Learning, ICML}

\vskip 0.3in
]



\printAffiliationsAndNotice{\icmlEqualContribution}  % leave blank if no need to mention equal contribution


\begin{abstract}
Adversarial training ({\AT}) is a widely recognized defense mechanism to gain the robustness of deep neural networks against  adversarial attacks. It  %
is built on min-max optimization (MMO), where the minimizer  (\emph{i.e.}, defender) seeks a robust model to minimize the worst-case training loss in the presence of adversarial examples crafted by the maximizer (\emph{i.e.}, attacker).
However,  the conventional MMO method  makes {\AT} hard to scale.
Thus,    \textit{{\FAT}}   \cite{Wong2020Fast} and other recent algorithms 
attempt to simplify MMO by replacing its maximization step with the single gradient sign-based attack generation step.
Although easy to implement, {\FAT} lacks theoretical guarantees, and its empirical performance is  unsatisfactory due to the issue of robust catastrophic overfitting when training with strong adversaries.
In this paper, we advance {\FAT} from the fresh perspective of bi-level optimization (BLO).
We first show that the commonly-used {\FAT} is equivalent to using a
stochastic gradient algorithm
to solve a linearized BLO problem involving a sign operation. However, the discrete nature of the sign operation makes it difficult to understand the algorithm performance. Inspired by BLO, we design and analyze a new set of robust training algorithms termed \textbf{Fast} \textbf{B}i-level \textbf{{\AT}} (\FBAT), which effectively defends sign-based projected gradient descent (PGD) attacks without using any gradient sign method or   explicit robust regularization. 
In practice, we show our method yields substantial robustness improvements over baselines across multiple models and datasets. 
\RV{Codes
are available at \url{https://github.com/OPTML-Group/Fast-BAT}.}
\end{abstract}

\begin{table*}[htb]
\begin{center}
\vspace*{-2.2mm}
\caption{\footnotesize{Performance overview of proposed {\FBAT} vs. the baselines {\FAT} \citep{Wong2020Fast} and {\FATGA} \citep{andriushchenko2020understanding} on CIFAR-10, CIFAR-100 and Tiny-ImageNet with PreActResNet-18. All methods are robustly {trained} under two perturbation budgets $\epsilon = 8/255$ and $16/255$  over 20 epochs. We use the early-stopping policy \citep{rice2020overfitting} to report the model with best robustness for each method.
The evaluation metrics include robust accuracy (RA) against PGD-50-10 attacks (50-step PGD attack with 10 restarts) \citep{madry2018towards} at $\epsilon = 8/255$ and $16/255$ (the test-time $\epsilon$ is the  \textit{same} as the train-time),
RA against  AutoAttack (AA) \citep{croce2020reliable} at $\epsilon = 8/255$ and $16/255$, and computation time (per epoch). The result $a${\tiny{$\pm b$}} represents mean $a$ and standard deviation $b$ over {10} random trials. All experiments are run on a single GeForce RTX 3090 GPU. 
}
} 
\label{table: motivation_intro_overview}
\vspace*{0.1in}
\begin{threeparttable}
\resizebox{0.8\textwidth}{!}{
\begin{tabular}{c|c|c|c|c|c|c|c}
\toprule[1pt]
\midrule
\multicolumn{8}{c}{{CIFAR-10, PreActResNet-18}} \\ \midrule
Method 
& \begin{tabular}[c]{@{}c@{}}
SA (\%) \\
($\epsilon = 8/255$)
\end{tabular}
& \begin{tabular}[c]{@{}c@{}}
RA-PGD (\%) \\
($\epsilon = 8/255$)
\end{tabular} & 
\begin{tabular}[c]{@{}c@{}}
RA-AA (\%) \\
($\epsilon = 8/255$)
\end{tabular}
& \begin{tabular}[c]{@{}c@{}}
SA (\%) \\
($\epsilon = 16/255$)
\end{tabular} 
& 
\begin{tabular}[c]{@{}c@{}}
RA-PGD (\%) \\ ($\epsilon = 16/255$)
\end{tabular}
& 
\begin{tabular}[c]{@{}c@{}}
RA-AA (\%) \\ ($\epsilon = 16/255$)
\end{tabular}
 & \begin{tabular}[c]{@{}c@{}}
Time \\ (s/epoch)
%Training time\\per epoch (s)
\end{tabular} \\ \midrule
{\FAT} 
& \textbf{82.39}\footnotesize{$\pm 0.14$} 
& 45.49\footnotesize{$\pm 0.21$} 
& 41.87\footnotesize{$\pm 0.15$} 
& 44.15\footnotesize{$\pm 7.27$} 
& 21.83\footnotesize{$\pm 1.32$} 
& 12.49\footnotesize{$\pm 0.33$} 
& 23.1 \\

{\FATGA} 
& 79.71\footnotesize{$\pm 0.24$} 
& 47.27\footnotesize{$\pm 0.22$} 
& 43.24\footnotesize{$\pm 0.27$} 
& 58.29\footnotesize{$\pm 1.32$} 
& 26.01\footnotesize{$\pm 0.16$} 
& 17.97\footnotesize{$\pm 0.33$} 
& 75.3\\

\rowcolor{Gray} \textbf{\FBAT} 
& 79.97\footnotesize{$\pm 0.12$} 
& \textbf{48.83}\footnotesize{$\pm 0.17$} 
& \textbf{45.19}\footnotesize{$\pm 0.12$} 
& \textbf{68.16}\footnotesize{$\pm 0.25$} 
& \textbf{27.69}\footnotesize{$\pm 0.16$} 
& \textbf{18.79}\footnotesize{$\pm 0.24$} 
& 61.4 \\\midrule

\multicolumn{8}{c}{{CIFAR-100, PreActResNet-18}} \\ \midrule
{\FAT} 
& \textbf{52.62} \footnotesize{$\pm 0.18$} 
& 24.66\footnotesize{$\pm 0.21$} 
& 21.72\footnotesize{$\pm 0.17$} 
& 21.32\footnotesize{$\pm 3.27$} 
& 8.62\footnotesize{$\pm 1.03$} 
& 6.22\footnotesize{$\pm 0.61$} 
& 23.8 \\

{\FATGA} 
& 50.06\footnotesize{$\pm 0.27$} 
& 24.97\footnotesize{$\pm 0.23$} 
& 21.82\footnotesize{$\pm 0.21$} 
& 32.51\footnotesize{$\pm 1.27$} 
& 12.27\footnotesize{$\pm 0.36$} 
&  9.43\footnotesize{$\pm 0.19$} 
& 77.1 \\

\rowcolor{Gray} \textbf{\FBAT} 
& 50.19\footnotesize{$\pm 0.21$} 
& \textbf{26.49}\footnotesize{$\pm 0.20$} 
& \textbf{23.97}\footnotesize{$\pm 0.15$} 
& \textbf{39.29}\footnotesize{$\pm 0.53$} 
& \textbf{13.97}\footnotesize{$\pm 0.17$} 
& \textbf{11.32}\footnotesize{$\pm 0.22$} 
& 61.6 \\\midrule

\multicolumn{8}{c}{{Tiny-ImageNet, PreActResNet-18}} \\ \midrule
{\FAT} 
& 41.37\footnotesize{$\pm 3.08$} 
& 17.05\footnotesize{$\pm 3.25$} 
& 12.31\footnotesize{$\pm 2.73$} 
& 31.38\footnotesize{$\pm 0.19$} 
&  5.42\footnotesize{$\pm 2.17$} 
&  3.13\footnotesize{$\pm 0.24$} 
& 284.6 \\

{\FATGA} 
& 45.52\footnotesize{$\pm 0.24$} 
& 20.39\footnotesize{$\pm 0.19$} 
& 16.25\footnotesize{$\pm 0.17$} 
& 29.17\footnotesize{$\pm 0.32$} 
&  6.79\footnotesize{$\pm 0.27$} 
&  4.27\footnotesize{$\pm 0.15$} 
& 592.7 \\

\rowcolor{Gray} \textbf{\FBAT} 
& \textbf{45.80}\footnotesize{$\pm 0.22$} 
& \textbf{21.97}\footnotesize{$\pm 0.21$} 
& \textbf{17.64}\footnotesize{$\pm 0.15$} 
& \textbf{33.78}\footnotesize{$\pm 0.23$} 
& \textbf{8.83}\footnotesize{$\pm 0.22$} 
& \textbf{5.52}\footnotesize{$\pm 0.14$} 
& 572.4 \\\midrule
\bottomrule[1pt]
\end{tabular}}
\end{threeparttable}
\vspace*{-6mm}
\end{center}
\end{table*}

\section{Introduction}
Given the fact that machine learning (ML) models can be easily fooled by tiny adversarial perturbations (also known as adversarial attacks) on the input \citep{Goodfellow2014explaining,carlini2017towards,papernot2016limitations}, training robust deep neural networks is now a major focus in research. 
Nearly all existing effective defense mechanisms \citep{madry2018towards,zhang2019theoretically,shafahi2019adversarial,Wong2020Fast,zhang2019you,athalye2018obfuscated} are built on the adversarial training ({\AT})  recipe, introduced by \citep{szegedy2014intriguing} and later formalized in \citep{madry2018towards} using \emph{min-max optimization} (MMO), 
where a minimizer~(\emph{i.e.} defender) seeks to update model parameters against a maximizer~(\emph{i.e.} attacker) that aims to increase the training loss by perturbing the training data.

 Despite the effectiveness of the {\AT}-type defenses in various application domains, the min-max nature makes them difficult to scale, because of the \textit{multiple} maximization steps required by the iterative attack generator at each training step. 
The resulting  prohibitive computation cost makes  {\AT}  not suitable in practical settings.   
For example, \cite{xie2019feature} used $128$ GPUs to run {\AT}   on ImageNet. 
Thereby, how to speed up {\AT} without losing  accuracy and robustness is now a \textit{grand challenge} for adversarial defense \citep{Wong2020Fast}.
 
Recently, \citep{Wong2020Fast,shafahi2019adversarial,zhang2019you,andriushchenko2020understanding} attempted 
to develop computationally-efficient alternatives of  {\AT}, that is, the `fast' versions of {\AT}.
To the best of our knowledge,  {\FAT} \citep{Wong2020Fast} and   {\FAT} with {g}radient {a}lignment (GA) regularization, termed {\FATGA} \citep{andriushchenko2020understanding},  are the two state-of-the-art (SOTA) `fast' versions of {\AT}, since they achieve the most significant reduction in computational complexity and preserve accuracy and robustness to a large extent.
{In particular, the inner-maximization in {\FAT} \citep{Wong2020Fast} only calls for single-step attack generation. However, different from the direct application of  fast gradient sign method \cite{Goodfellow2015explaining}, the empirical success of {\FAT} also relies on a series of heuristics-based strategies, \textit{e.g.}, using  large attack step size, cyclic learning rate schedule, and mixed-precision training.}
Yet, {\FAT} suffers from two main {issues}: \textit{(i)} lack of stability, \textit{i.e.},  it has a large variance in  performance \citep{li2020towards}, and \textit{(ii)} catastrophic overfitting, \textit{i.e.}, when training with strong adversaries, the robustness of the resulting model can drop significantly. 
To alleviate these problems, \citet{andriushchenko2020understanding} proposed {\FATGA} by penalizing   {\FAT} using  an explicit GA regularization. However, we will show that {\FATGA} encounters another problem: \textit{(iii)} It improves   robust accuracy (RA) at the cost of a sharp drop in standard accuracy (SA), leading to a poor accuracy-robustness tradeoff for large attack budget (\textit{e.g.}, $\epsilon=16/255$ in Table\,\ref{table: motivation_intro_overview}). Moreover, \textit{(iv)} there has been no theoretical guarantee for the optimization algorithms used in {\FAT} and {\FATGA}.
Given the limitations \textit{(i)}-
{\textit{(iv)}}, we ask: 

{
\vspace*{-2mm}
\begin{center}
   \textit{How to design a
   `fast'   {\AT} with {improved stability}, {mitigated catastrophic overfitting}, {enhanced RA-SA tradeoff}, {and some theoretical guarantees}?}
\end{center}
\vspace*{-2mm}
}

To address the above question, we formulate the {\AT} problem as a unified \textbf{bi-level optimization (BLO)} problem \citep{dempe2002foundations}.
In the new formulation, the attack generation is cast as a constrained \textit{lower-level} optimization problem, while the defense serves as the \textit{upper-level} optimization problem. To the best of our knowledge, this is the first work that makes a rigorous connection between adversarial defense and BLO. 
Technically, we show that {\FAT} can be \textit{interpreted} as BLO with linearized  lower-level problems. Delving into the linearization of BLO, we propose a novel, theoretically-grounded `fast' AT framework, termed \textbf{fast} \textbf{b}i-level \textbf{{\AT}} ({{\FBAT}}). Practically,
Table\,\ref{table: motivation_intro_overview} highlights some achieved improvements   over {\FAT} and {\FATGA}:
When a stronger train-time attack (\textit{i.e.},  $\epsilon = 16/255$) is adopted, {\FAT}  suffers from a large degradation of RA and SA, together with a very high variances. Although {\FATGA} outperforms {\FAT}, it still incurs a significant SA loss at $\epsilon = 16/255$. In contrast, {\FBAT} produces the best robustness with a more graceful SA-RA tradeoff: As {\FBAT} brings an improvement of over $1.5\%$ on RA in all experimental settings, its SA still remains at a high level, \emph{e.g.} a significant improvement on SA by $9.9\%$ and $6.7\%$ at $\epsilon = 16/255$ for CIFAR-10 and CIFAR-100.

\vspace{-3mm}
\paragraph{Contributions.}
We summarize our contributions below.

\ding{172} (Formulation-wise) We propose a unified BLO-based formulation for the robust training problem. 
Within this formulation, we show the conventional {\FAT} method is solving a low-level linearized BLO problem, rather than   the vanilla min-max problem.
This key observation not only provides a new interpretation of {\FAT}, but most importantly, also explains why {\FAT} is difficult to possess strong theoretical guarantees.

\ding{173} (Methodology-wise) We propose the new {\FBAT} algorithm   based on our new understanding of {\FAT}. The key enabling technique is to introduce a new \emph{smooth} lower-level objective of BLO for robust training. In contrast to MMO, BLO adopts a different optimization routine that requires implicit gradient ({IG}) computation. We derive the closed-form of IG   for {\FBAT}.

\ding{174} (Experiment-wise) We made a comprehensive experimental study to demonstrate the effectiveness of  {\FBAT} over SOTA baselines across datasets and model types.  Besides its merit in robustness enhancement, we also show  its improved stability, lifted accuracy-robustness trade-off, and mitigated catastrophic overfitting.

\section{Related work}

\paragraph{Adversarial attack.}
Adversarial attacks are techniques to generate malicious perturbations that are imperceptible to humans but can mislead the ML models \citep{Goodfellow2014explaining,carlini2017towards, croce2020reliable,xu2019structured,athalye2018synthesizing}.
The adversarial attack has become a major approach to evaluate the robustness of deep neural networks and thus, help build safe artificial intelligence in many high-stakes applications such as autonomous driving~\citep{deng2020analysis,kumar2020black} and  surveillance~\citep{thys2019fooling, xu2020adversarial}.

\paragraph{Adversarial defense and robust training at scale.}
Our work falls into the  category of robust training,  which was
mostly built on MMO. 
For example,   \cite{madry2018towards} established the framework of {\AT} for the first time, always recognized as one of the most powerful defenses %
\citep{athalye2018obfuscated}.
Extended from {\AT}, TRADES \citep{zhang2019theoretically} sought the optimal balance between robustness and  generalization ability. Further, {\AT}-type defense has been generalized to  the semi-/self-supervised settings
\citep{carmon2019unlabeled,chenliu2020cvpr}
and integrated \ref{eq: prob_AT} with certified defense techniques such as randomized smoothing~\citep{salman2019provably}.

Despite the effectiveness of {\AT} and its variants, 
how to speed up {\AT} without losing performance remains an open question. 
Some recent works attempted to impose algorithmic simplifications to {\AT}, leading to
    \textit{fast but approximate} {\AT} algorithms, such as `free' AT \citep{shafahi2019adversarial},  you only propagate once (YOPO) \citep{zhang2019you},   {\FAT} \citep{Wong2020Fast}, and {\FAT} regularized by gradient alignment (termed {\FATGA}) \citep{andriushchenko2020understanding}. 
In particular, {\FAT} and {\FATGA} are the baselines most relevant to ours due to their low computational complexity. However, their defense performance is still unsatisfactory. For example, {\FAT} has  poor   training stability \citep{li2020towards} and   suffers  catastrophic overfitting when facing strong attacks  \citep{andriushchenko2020understanding}. In contrast,  {\FATGA} yields improved robustness but has a poor accuracy-robustness tradeoff (see Table\,\ref{table: motivation_intro_overview}).

\paragraph{Bi-level optimization (BLO).}
BLO is a unified hierarchical learning framework, where the objective and variables of the upper-level problem depend on the lower-level problems.  The BLO problem in its most generic form is very challenging, and thus, the design of algorithms and theory for BLO focuses on special cases
 \citep{vicente1994descent,white1993penalty,gould2016differentiating,ghadimi2018approximation,khanduri2021momentum,ji2020bilevel, hong2020two}.
In practice, BLO has been successfully applied to meta-learning \citep{rajeswaran2019meta}, data poisoning attack design \citep{huang2020metapoison}, and reinforcement learning \citep{chen2019research}.
{Recently, references \cite{zuo2021adversarial} formulated the AT training problem using a Stackelberg Game, which is a special case of bi-level optimization. Different from ours, the goal of \cite{zuo2021adversarial} is to improve model generalization in natural language processing tasks, rather than improving model  robustness in vision tasks. Further, note that in our formulation, we use a generic lower-level problem to model the adversarial learning problem, while in \cite{zuo2021adversarial} a multi-step unrolling of certain update mapping is used for this purpose. Importantly, with our formulation, it is possible to provide theoretical analysis for the resulting algorithms,  under some conditions about the lower-level problem.
%and such analysis is missing in \cite{zuo2021adversarial}.
}
As will be evident later, the existing BLO approach is not directly applied to adversarial defense due to the presence of the \textit{constrained} non-convex lower-level problem (for attack generation). 
To the best of our knowledge, our work makes a rigorous connection between adversarial defense and BLO for the first time.

\section{A Bi-Level Optimization View on {\FAT}}

\mycomment{
In this section, we first provide a brief background on 
the formulation and the algorithm backbone   of 
{\FAT}. We then introduce the setup of BLO   and explore its connection to adversarial robustness. 
}

\paragraph{Preliminaries on {\AT} and {\FAT}.}
{Let us consider the following standard min-max formulation for the robust adversarial training problem \citep{madry2018towards}}%

{\vspace*{-5mm} \small\begin{align}
    \begin{array}{l}
       \displaystyle  \minimize_{\btheta} ~  \mathbb E_{(\mathbf x, y) \in \mathcal D} \left [
       \maximize_{\boldsymbol \delta \in \mathcal C}
       \ell_{\mathrm{tr}}(\btheta, \mathbf x + \bdelta, y)
       \right ], 
    \end{array}
    \label{eq: prob_AT} 
\end{align}}%
\vspace*{-4mm}

where $\btheta \in \mathbb R^n$ denotes model parameters; $\mathcal D$ is the training set consisting  (a finite number) of labeled data pairs with feature $\mathbf x$ and label $y$; $\bdelta \in \mathbb R^d $ represents   adversarial perturbations subject to the perturbation constraint   $\mathcal C$, {\textit{e.g.}}, $ \mathcal{C} = \{  \bdelta \, | \, \| \bdelta \|_\infty  \leq \epsilon, \bdelta \in [\mathbf 0, \mathbf 1] \}$ for $\epsilon$-toleration $\ell_\infty$-norm   constrained attack (normalized to $[\mathbf 0, \mathbf 1]$); $\ell_{\mathrm{tr}}(\cdot)$ is the training loss;
% Using the above notation,
$(\mathbf{x} + \bdelta)$ represents an adversarial example. %

The standard solver for problem \eqref{eq: prob_AT} is known as {\AT} \citep{madry2018towards}. However,  it has to call an \textit{iterative} optimization method (\textit{e.g.}, $K$-step PGD attack)   to solve the inner maximization problem of  \eqref{eq: prob_AT}, which is computationally expensive. 
To improve its scalability, the {\FAT} algorithm \citep{Wong2020Fast} was proposed to take a \textit{single-step} PGD attack for inner maximization. The {algorithm backbone} of {\FAT} is summarized below.

\paragraph{{\FAT} algorithm}
Let $\btheta_{t}$ be       parameters at  iteration $t$. The $(t+1)$th   iteration becomes \cite{Wong2020Fast}:

\noindent $\bullet$ \textit{Inner maximization by $1$-step PGD:}
    {\vspace*{-1mm} \small\begin{align*}
        \bdelta \xleftarrow{} \mathcal P_{\mathcal C} \left   ( \bdelta_0 + \alpha \cdot  \mathrm{sign} \left ( \nabla_{\bdelta} \ell_{\mathrm{tr}}(\btheta_t, \mathbf x+\bdelta_0, y) \right  ) \right  ),
    \end{align*}}%
    \vspace*{-7mm}

\noindent where $\mathcal P_{\mathcal C} (\mathbf a)$ denotes the projection operation that projects the point $\mathbf a$ onto $\mathcal C$, \textit{i.e.},  $\mathcal P_{\mathcal C} (\mathbf z) = \argmin_{\bdelta \in \mathcal C} \| \bdelta - \mathbf z \|_2^2$, $\bdelta_0$ is a random uniform initialization within $[\mathbf 0, \mathbf 1]$, $\alpha > 0$ is a proper learning rate (\textit{e.g.}, $1.25 \epsilon$), and $\mathrm{sign}(\cdot)$ is the element-wise sign operation.

\noindent $\bullet$ \textit{Outer minimization for model training:} This can be conducted by any standard optimizer, \textit{e.g.}, 
    
    {\vspace*{-7mm} \small \begin{align*}
        \btheta_{t+1} \xleftarrow{} \btheta_{t} -  \beta   \nabla_{\btheta} \ell_{\mathrm{tr}}(\btheta_t, \mathbf x + \bdelta, y)
    \end{align*}}%
    \vspace*{-7mm}
    
where $\beta > 0$ is a proper learning rate (\textit{e.g.}, cyclic learning rate), and $\bdelta$ is provided from the inner maximization step. 

A few remarks about {\FAT} are given below.

\ding{172} Roughly speaking, {\FAT} is a simplification of {\AT} using $1$-step PGD for inner maximization. However, it is not entirely clear which problem \FAT~is actually solving. If we take a closer look at the algorithm, we will see that the inner update only changes to the {\it initial} $\boldsymbol \delta_0$, not the most recent  $\boldsymbol \delta$. Clearly, this scheme is fundamentally {\it different} from the classical gradient descent-ascent algorithm for min-max optimization  \cite{razaviyayn2020nonconvex}, which alternatively updates the inner and outer variables. 
\emph{Therefore, it remains elusive  if \FAT~is   solving the original   problem \eqref{eq: prob_AT}}.

\iffalse
\ding{173} As  shown in \citep{Wong2020Fast},  the successful implementation of {\FAT} needs sophisticated hyperparameter tuning for   $\alpha$, $\bdelta_0$, and $\beta$. 
\fi 

\ding{173} \citet{andriushchenko2020understanding} demonstrated that {\FAT} could lead to catastrophic overfitting
when using strong adversaries for training. However, there was no grounded theory to justify the pros and cons of {\FAT}. We will show that BLO provides a promising solution.

\paragraph{BLO: Towards a unified formulation of robust training.}
{BLO (bi-level optimization) is an optimization problem that involves two levels of optimization tasks, where the {\it lower-level} task  is
nested inside the {\it upper-level} task. More precisely, it has the following generic form:

\vspace*{-4mm}
{\small\begin{equation}
\begin{array}{cl}
\label{eq:blo}
    \displaystyle \min_{\mathbf {u}\in \mathcal{U}} &\quad f({\mathbf u}, {\mathbf v}^*({\mathbf u}))\\
  \text{s.t.}  & \quad {\mathbf v}^*({\mathbf u}) = \arg\min_{\mathbf v\in \mathcal{V}}\;  g(\mathbf v, \mathbf u)   
\end{array}
\end{equation}}%
\vspace*{-5mm}

where $\mathcal{U}$ and $\mathcal{V}$ are the feasible sets for the variables $\mathbf u$ and $\mathbf v$, respectively; $f(\cdot)$ and $g(\cdot)$ are the upper- and the lower-level objective functions, respectively. 
Intuitively, the BLO \eqref{eq:blo} can be used to formulate the  adversarial training problem as the latter also involves two problems, one nested in the other. Importantly, it is more powerful than the min-max formulation \eqref{eq: prob_AT} as it allows the two problems to have {\it different} objective functions. This flexibility provided by BLO is the key to the generality of our proposed framework. 

To make the above intuition precise,
we use the upper-level problem to model the training loss minimization, while the lower-level problem to model the attack generation process, and consider the following BLO problem: 
}%

{\vspace*{-7mm}		\small	\begin{align}
\begin{array}{cl}
\displaystyle \min_{\boldsymbol \theta }         & \mathbb E_{(\mathbf x, y) \in \mathcal D} [ \ell_{\mathrm{tr}}(\boldsymbol \theta, \mathbf x +  \boldsymbol \delta^*(\boldsymbol \theta; \mathbf x, y); y) ] \\
\text{s.t.}  & \boldsymbol \delta^*(\boldsymbol \theta; \mathbf x, y) = \displaystyle \argmin_{ \boldsymbol \delta \in \mathcal C}  \ell_{\mathrm{atk}}( \boldsymbol \theta, \boldsymbol \delta; \mathbf x, y),
\end{array}
\label{eq: prob_biLevel}
\end{align}}%
\vspace*{-7mm}

where the training loss function $\ell_{\rm tr}(\cdot)$ has been defined in \eqref{eq: prob_AT}, and $\ell_{\mathrm{atk}}(\cdot)$ denotes an attack objective.
For the notation simplicity, in the subsequent discussion, we will not indicate the dependency of the functions $\ell_{\mathrm{tr}}$, $\ell_\mathrm{atk}$, and $\bdelta^*$ with respect to (w.r.t.) the data samples $(\mathbf{x},y)$.
The  formulation \eqref{eq: prob_biLevel} has two key differences from \eqref{eq: prob_AT}:

\ding{172} When we choose $\ell_{\rm atk} = - \ell_{\rm tr}$, problem \eqref{eq: prob_biLevel} becomes equivalent to the min-max formulation \eqref{eq: prob_AT}. It follows that the BLO is suitable to formulate the adversarial training problem. Moreover, we will see shortly that the flexibility provided by choosing the lower-level objective {\it independently} of the upper-level one enables us to interpret {\FAT} as solving a certain special form of the BLO problem. Note that prior to our work, it was not entirely clear what is the problem that {\FAT} is trying to solve.

\ding{173} BLO calls a different optimization routine from those to solve the original min-max   problem \eqref{eq: prob_AT}. As will be evident later, even if we set  $\ell_{\mathrm{atk}} = - \ell_{\mathrm{tr}}$ in \eqref{eq: prob_biLevel}, the  BLO-enabled solver  does not simplify to {\FAT} ({see more details in Appendix\,\ref{app: proof_problem2_problem1}}). %
This is because for a given data sample $(\mathbf{x},y)$, the   gradient for the upper-level problem of \eqref{eq: prob_biLevel} yields:

\vspace*{-6mm}
{\small
\begin{equation}
\displaystyle
\resizebox{.42\textwidth}{!}{$
\text{\resizebox{.13\textwidth}{!}{$\hspace*{-3.5mm} \frac{d \ell_{\mathrm{tr}}(\boldsymbol \theta, \boldsymbol \delta^*(\boldsymbol \theta))}{d \btheta }$}}
    =
    \nabla_{\btheta} \ell_{\mathrm{tr}}(\boldsymbol \theta, \boldsymbol \delta^*(\btheta))
    + \underbrace{ \frac{{d} \boldsymbol \delta^*(\boldsymbol \theta)^\top}{ {d} \boldsymbol \theta}  }_\text{{\IG}}
    \nabla_{\bdelta} \ell_{\mathrm{tr}}(\btheta, \bdelta^*(\btheta))
    $}
    , \hspace*{-3mm} \label{eq: GD_upper}
\end{equation}}%
\vspace*{-7mm}

\noindent where the superscript $\top$ denotes the transpose operation, 
and $\nabla_{\btheta} \ell_{\mathrm{tr}}(\boldsymbol \theta, \boldsymbol \delta^*(\btheta))$ denotes the partial derivative w.r.t. the first input argument $\btheta$; and  $ \frac{d \boldsymbol \delta^*(\boldsymbol \theta)^\top}{ d \boldsymbol \theta} \in \mathbb R^{n \times d}$, if exists,  is referred to  as \textit{{implicit gradient} ({\IG})} because  it is defined through an implicitly constrained optimization problem $\min_{ \boldsymbol \delta \in \mathcal C}  \ell_{\mathrm{atk}}$.  
The dependence on {\IG} is a `fingerprint' of   BLO   \eqref{eq: prob_AT} in contrast to   {\AT} or {\FAT}. 

\paragraph{BLO-enabled interpretation of {\FAT}.}
Next, we demonstrate how {\FAT} relates to BLO. {\FAT} can be interpreted as {an approximated stochastic gradient algorithm for solving the following}  \textbf{lower-level linearized BLO}. That is to say, {\FAT} is not solving the original min-max problem \eqref{eq: prob_AT}, but its linearized version below:
\begin{tcolorbox}
\vspace*{-5mm}
{\small \begin{align}\label{eq: prob_biLevel_sign_lin} 
\raisetag{10mm}
\hspace*{-3.5mm}
\text{\resizebox{7.2cm}{!}{$
\begin{array}{cl}
\hspace*{-5mm} 
\min_{\boldsymbol \theta } &
\hspace*{-2mm}
\mathbb E_{(\mathbf x, y) \in \mathcal D} [ \ell_{\mathrm{tr}}(\boldsymbol \theta,  \boldsymbol \delta^*(\boldsymbol \theta)) ] \\
\hspace{-5mm}
\text{s.t.}& 
\hspace*{-3mm}
\boldsymbol \delta^*(\boldsymbol \theta) \hspace*{-1mm} 
= 
\hspace*{-1mm}
\argmin_{ \boldsymbol \delta \in \mathcal C } ~ [  (\bdelta - \mathbf z)^\top \mathrm{sign}(\nabla_{\bdelta = \mathbf z}\ell_{\mathrm{atk}}( \btheta, \bdelta) ) + (\lambda/2) \| \boldsymbol \delta -\mathbf z \|_2^2 ] ,
\hspace*{-9mm}
\end{array}
$}}
\end{align}}%
\vspace*{-6mm}
\end{tcolorbox}

where $\mathbf z = \bdelta_0$ and $\lambda = 1/\alpha$. Our justification for the above interpretation is elaborated  on below.  

\ding{172} The simplified lower-level problem of \eqref{eq: prob_biLevel_sign_lin} leads to the \textbf{closed-form} solution:

{
\vspace*{-6mm}
\small \begin{equation}
\text{\resizebox{.42\textwidth}{!}{$
\begin{aligned}
\hspace*{-3.5mm} \bdelta^*(\btheta) = & \displaystyle \argmin_{ \boldsymbol \delta \in \mathcal C } \,
(\lambda / 2) \| \boldsymbol \delta -\mathbf z + (1/\lambda) \mathrm{sign}(\nabla_{\bdelta = \mathbf z}\ell_{\mathrm{atk}}( \btheta, \bdelta) ) \|_2^2 \\
= & \mathcal P_{\mathcal C} \left (
\mathbf z - (1/\lambda) \mathrm{sign}(\nabla_{\bdelta = \mathbf z }\ell_{\mathrm{atk}}( \btheta, \bdelta) ) \right ), \hspace*{-5mm}
\end{aligned}$}}
\label{eq: sol_sign_linear}
\end{equation}
}%
\vspace*{-4mm}

which is exactly given by the $1$-step PGD attack with initialization $\mathbf z$ and learning rate $(1/\lambda)$.
{In the linearization used in \eqref{eq: prob_biLevel_sign_lin}, a quadratic regularization term (with regularization parameter $\lambda$) is introduced to ensure the strong convexity of  the lower-level objective within the constraint set $\boldsymbol \delta \in \mathcal C$ so as to achieve the unique minimizer \eqref{eq: sol_sign_linear}.
Note that imposing such a strongly convex regularizer is also commonly used to stabilize the convergence of MMO (min-max optimization) and BLO ~\citep{qian2019robust,hong2020two}.}
If we set $\mathbf z = \bdelta_0$ and $\lambda = 1/ \alpha$, \eqref{eq: sol_sign_linear}  precisely depicts  the inner maximization step used in {\FAT}.

\ding{173} By substituting \eqref{eq: sol_sign_linear}
into the upper-level problem of \eqref{eq: prob_biLevel_sign_lin}, 
we can then use \eqref{eq: GD_upper} to compute the stochastic gradients of the upper-level problem. If the stochastic gradient can be precisely computed, we can update the model parameters $\btheta$ using 
SGD based on \eqref{eq: GD_upper}. That is, 
$
 \scriptstyle{\btheta \xleftarrow{} \btheta - \beta} \textstyle{\frac{d \ell_{\mathrm{tr}}(\boldsymbol \theta_t, \boldsymbol \delta^*(\boldsymbol \theta))}{d \btheta }}
$ (with learning rate $\beta$).
However, generally speaking, the {\IG} function  
$ \frac{d \boldsymbol \delta^*(\boldsymbol \theta)^\top}{ d \boldsymbol \theta} $ involved in \eqref{eq: GD_upper}
may not be differentiable, and even it is, the computation may not be easy. For our case, $\boldsymbol \delta^*(\boldsymbol \theta)$ expressed in \eqref{eq: sol_sign_linear} involves both a projection and a sign operation, which can be particularly difficult to compute. To proceed, 
let us make the following approximations. 
We assume that the chain rule of the derivative of  $\boldsymbol \delta^*(\boldsymbol \theta)$ holds w.r.t. $\boldsymbol \theta$, implying the differentiability of the projection operation and the sign operation.
Then, based on the closed-form of $\boldsymbol \delta^*(\boldsymbol \theta)$ in \eqref{eq: sol_sign_linear}, {\IG} is approximately equal to \mathHigh{\textstyle\frac{d \bdelta^*(\btheta)^\top}{d \btheta}\scriptstyle=\mathbf 0}, where we use two {facts}:
(1) The linearization point $\mathbf z$ is   independent of $\btheta$, \textit{i.e.} $\mathbf z = \bdelta_0$; And (2) $\frac{d \mathrm{sign}(\cdot)}{ d \boldsymbol \theta} = \mathbf 0$ holds almost everywhere. 
Clearly, the use of the gradient sign method simplifies the {\IG} computation. Further,
the upper-level gradient is approximated by
$
\frac{d \ell_{\mathrm{tr}}(\boldsymbol \theta, \boldsymbol \delta^*(\boldsymbol \theta))}{d \btheta }
 \scriptstyle
 \approx 
    \nabla_{\btheta} \ell_{\mathrm{tr}}(\boldsymbol \theta, \boldsymbol \delta^*(\btheta)) \Def \tilde {\mathbf h} (\boldsymbol \theta)  
$, and the upper-level optimization of problem \eqref{eq: prob_biLevel_sign_lin} becomes
$\scriptstyle \btheta \xleftarrow{} \btheta - \beta   \tilde {\mathbf h} (\boldsymbol \theta)  $, which is the same as the outer minimization step in {\FAT}. 
In a nutshell, the BLO solver of problem \eqref{eq: prob_biLevel_sign_lin}, which   calls the IG computation based on \eqref{eq: sol_sign_linear}, eventually reduces to the {\FAT} algorithm.

{The aforementioned analysis shows that {\FAT} can be viewed as using an approximated stochastic gradient algorithm to solve the linearized BLO \eqref{eq: prob_biLevel_sign_lin}, with the linearization point $\mathbf z$ and the regularization parameter $\lambda$ set as  $\mathbf z = \bdelta_0$ and $\lambda = 1/\alpha$. However, since a series of approximations have been used when arriving at the approximated gradient $\widetilde{\mathbf h}(\boldsymbol \theta)$ used by \FAT, it is no longer clear if %
the resulting algorithm can still sufficiently reduce the objective function of the upper-level problem. Additionally, based on the fact that the lower-level problem of \eqref{eq: prob_biLevel_sign_lin} involves the \textit{discrete} sign operator, it is unlikely (if not impossible) that any approximated stochastic gradient-based algorithms developed for it can possess any strong theoretical guarantees. }

\vspace*{-2mm}
\section{{\FBAT}: Advancing {\FAT} by BLO}
\mycomment{
In this section, we leverage the BLO view \eqref{eq: prob_biLevel_sign_lin} to improve  {\FAT} and propose an advanced {\FAT} framework, termed      {\FBAT}.
}

\paragraph{{\FBAT} and rationale.}
The key take-away   
 from \eqref{eq: prob_biLevel_sign_lin}
is that the conventional {\FAT} adopts 
the \textit{sign of input gradient} to linearize the lower-level attack objective.  
However, a more natural  choice  is to use the   first-order Taylor expansion for   linearization. 
By doing so, problem \eqref{eq: prob_biLevel_sign_lin} can be  modified to the following form:

\vspace*{-2mm}
\begin{tcolorbox}
\vspace*{-5mm}
{\small \begin{align}\label{eq: prob_biLevel_lin} 
\raisetag{10.5mm}
\hspace*{-3.5mm}
\text{\resizebox{7.2cm}{!}{$
\begin{array}{cl}
\hspace*{-4mm} 
\min_{\boldsymbol \theta } &
\hspace*{-2mm}
\mathbb E_{(\mathbf x, y) \in \mathcal D} [ \ell_{\mathrm{tr}}(\boldsymbol \theta,  \boldsymbol \delta^*(\boldsymbol \theta)) ] \\
\hspace{-5mm}
\text{s.t.}& 
\hspace*{-3mm}
\boldsymbol \delta^*(\boldsymbol \theta) \hspace*{-1mm} 
= 
\hspace*{-1mm}
\argmin_{ \boldsymbol \delta \in \mathcal C } ~ [  (\bdelta - \mathbf z)^\top (\nabla_{\bdelta = \mathbf z}\ell_{\mathrm{atk}}( \btheta, \bdelta) ) + (\lambda/2) \| \boldsymbol \delta -\mathbf z \|_2^2 ] ,
\hspace*{-8mm}
\end{array}
$}}
\end{align}}%
\vspace*{-6mm}
\end{tcolorbox}

Similar to \eqref{eq: sol_sign_linear}, problem \eqref{eq: prob_biLevel_lin} can be solved as:

{\vspace*{-7mm} \small \begin{align}
    \bdelta^*(\btheta) =
    \mathcal P_{\mathcal C} \left (
\mathbf z - (1/\lambda) \nabla_{\bdelta = \mathbf z }\ell_{\mathrm{atk}}( \btheta, \bdelta)  \right ).
\label{eq: sol_linear}
\end{align}}%
\vspace*{-7mm}

\noindent In contrast to \eqref{eq: sol_sign_linear}, the {\IG}  associated with \eqref{eq: prob_biLevel_lin} cannot be approximated by zeros  since the gradient sign operation is not present in 
\eqref{eq: sol_linear}. 
To compute   {\IG}, the auto-differentiation (the chain rule)
can be applied to the closed-form of $\bdelta^*(\btheta)$. {This is also what has been used in \cite{zuo2021adversarial} that calls     projected gradient descent unrolling to solve  BLO problems.}
However, this will not give us an accurate and generalizable {\IG} solution since the projection operation $\mathcal P_{\mathcal{C}}$ is \textit{not} smooth everywhere {and thus, the use of chain rule does not yield a rigorous derivation}.  In the following subsection, we address the
  \textit{{\IG} challenge}  in a theoretically-grounded manner. 

\paragraph{{\IG} theory for {\FBAT}.}
The problem of {\FBAT} 
\eqref{eq: prob_biLevel_lin}
falls into a class of very challenging BLO problems, which requires \textit{constrained} lower-level optimization. 
The unconstrained case is easier to handle since one can apply the implicit function theory to   the stationary condition of the lower-level problem to obtain {\IG} \citep{hong2020two}. Yet,  in our case with \textit{constrained} problems, a stationary point could violate the constraints, and thus, the stationary condition becomes non-applicable.
Fortunately, in problem  
\eqref{eq: prob_biLevel_lin}, we are dealing with a special class of lower-level constraints -- \textit{linear constraints}. Let us rewrite the constraints below: 

\vspace*{-7mm}
{\small \begin{equation} \begin{split}
   \| \bdelta \|_\infty  \leq \epsilon, \bdelta \in [\mathbf{-x}, \mathbf{1-x}]  
  \Longleftrightarrow{} \mathbf B \bdelta \leq \mathbf b, 
\label{eq: lin_cons}
\end{split} \end{equation}}%
\vspace*{-7mm}

\noindent where {$\renewcommand*{\arraystretch}{0.7} \scriptstyle{ \mathbf B \Def} \begin{bmatrix}
    \scriptstyle{\mathbf I} \\
     \scriptstyle{- \mathbf I}
    \end{bmatrix}, ~ 
    \scriptstyle{\mathbf b \Def}  \begin{bmatrix}
     \scriptstyle{\min \{  \epsilon \mathbf 1, \mathbf 1 - \mathbf x \}}\\
     \scriptstyle{- \max\{ - \epsilon \mathbf 1, - \mathbf x \}}
    \end{bmatrix}$}.
By exploiting the above linearly constrained problem structure, we show that the {\IG} challenge associated with  \eqref{eq: prob_biLevel_lin} can be addressed via \textit{Karush–Kuhn–Tucker (KKT)} conditions. We present our main theoretical results (Proposition\,\ref{prop: 1} and Corollary\,\ref{eq: thr_IG}) below and refer readers to {Appendix\,\ref{app: IG_thr}} for detailed derivation. 

\begin{myprop}\label{prop: 1}
For the BLO problem \eqref{eq: prob_biLevel_lin},  let $g(\boldsymbol \theta, \boldsymbol \delta^*)$ denote its lower-level objective function
evaluated at $\boldsymbol \delta^*$ given $\boldsymbol \theta$, then the analytical form of {\IG} (implicit gradient) is given by

\vspace*{-6mm}
{\small \begin{align} 
\label{eq: IG_w_Hessian}
\raisetag{10mm}
\text{\resizebox{.47\textwidth}{!}{$
\renewcommand{\arraystretch}{2}
\begin{array}{cc}
    \hspace*{-5mm}
    &\hspace*{-50mm} \frac{d \bdelta^*(\btheta)^\top}{d \btheta}
    = -\nabla_{\btheta \bdelta} g(\btheta , \bdelta^\ast) \nabla_{\bdelta \bdelta} g(\btheta , \bdelta^\ast)^{-1} \\
    \hspace*{-5mm}
    & \hspace*{-0mm}+ \nabla_{\btheta \bdelta} g(\btheta , \bdelta^\ast) \nabla_{\bdelta \bdelta} g(\btheta , \bdelta^\ast)^{-1} \mathbf{B}_0^\top [\mathbf B_0 \nabla_{\bdelta \bdelta} g(\btheta, \bdelta^\ast)^{-1} \mathbf B_0^\top]^{-1} \mathbf B_0 \nabla_{\bdelta \bdelta} g(\btheta , \bdelta^\ast)^{-1},
\end{array}
$}}
\end{align}}%
\vspace*{-5mm}

\noindent where 
$\textstyle \bdelta^*$
is given  by  \eqref{eq: sol_linear} (the dependence on 
$\textstyle \btheta$
is omitted for ease of notation),  
$\textstyle \nabla_{\btheta \bdelta} g (\btheta , \bdelta^*) \in \mathbb R^{n \times d}$
denotes the second-order  partial derivatives evaluated at 
$\textstyle (\btheta, \bdelta^*)$, 
and 
$\textstyle \mathbf B_0$  
denotes  the sub-matrix of $\textstyle \mathbf B$
that only corresponds to    the {active constraints} at 
$\textstyle \bdelta^\ast$, 
\textit{i.e.}, those in 
$\textstyle \mathbf B \bdelta^\ast \leq \mathbf b$ 
satisfied with equality.
\end{myprop}

It is clear from \eqref{eq: IG_w_Hessian} that the computation of IG requires the second-order derivatives as well as matrix inversion. This is computationally intensive. 
Recall from  \eqref{eq: prob_biLevel_lin} that  the Hessian matrix $\nabla_{\bdelta \bdelta} g$  of 
 the lower-level objective function  is   given by
 $
 \nabla_{\bdelta \bdelta} g(\btheta , \bdelta^\ast) = \nabla_{\bdelta \bdelta} \ell_{\mathrm{atk}} +\lambda \mathbf I
 $. This inspires us to impose the Hessian-free approximation, \textit{i.e.}, $ \nabla_{\bdelta \bdelta} \ell_{\mathrm{atk}} = \mathbf 0$. 
The rationale behind the Hessian-free assumption     is that
     ReLU-based
      neural networks commonly
   lead to a piece-wise linear decision boundary w.r.t.
the inputs \citep{moosavi2019robustness,alfarra2020decision}, and thus, 
its   second-order derivative (Hessian) $\nabla_{\bdelta \bdelta}  \ell_{\mathrm{atk}}$  is close to zero.   
In Sec.\,\ref{sec: exp_robustness} and Appendix\,\ref{app: additional_result}, we will empirically  show that the Hessian-free assumption is reasonable for both ReLU and non-ReLU neural networks. 
Thus, the Hessian matrix is approximated as:

\vspace*{-5mm}
{\small \begin{align}\label{eq: Hessian_simplification_main}
    \nabla_{\bdelta \bdelta} g(\btheta , \bdelta^\ast(\btheta)) = \nabla_{\bdelta \bdelta} \ell_{\mathrm{atk}} +\lambda \mathbf I = \mathbf 0 + \lambda \mathbf I.
\end{align}}%
\vspace*{-7mm}

With \eqref{eq: Hessian_simplification_main}, we can simplify closed-form of IG as below:

\begin{mycor} 
\label{eq: thr_IG}
With the Hessian-free assumption, namely 
{$\nabla_{\bdelta \bdelta}\ell_{\mathrm{atk}} = \mathbf 0$}, 
the {\IG} (implicit gradient) of   \eqref{eq: prob_biLevel_lin} is %

{\vspace*{-3mm}
\small 
\begin{equation} 
   {\frac{d \bdelta^*(\btheta)^\top}{d \btheta}} 
    = - (1/\lambda)\nabla_{\btheta \bdelta} \ell_{\mathrm{atk}}(\btheta , \bdelta^\ast) \mathbf H_{\mathcal C},
    \label{eq: IG_fast_BAT} 
\end{equation} 
}%
\vspace*{-5mm}

\noindent where $\scriptstyle \mathbf H_{\mathcal C} \Def \begin{bmatrix}\scriptstyle
1_{p_1 < \delta_1^* < q_1  } \mathbf e_1 \cdots 1_{p_1 < \delta_d^* < q_d  } \mathbf e_d \end{bmatrix}\in \mathbb R^{d \times d}$ and  the function
$ 1_{p_i < \delta_i^* < q_i} \in \{ 0, 1\}$ denotes the indicator over the constraint of $\{ \delta_i \, | \, p_i < \delta_i^* < q_i \}$, which returns $1$ if the constraint is satisfied, $\delta_i^*$ denotes the $i$th entry of $\bdelta^*(\btheta)$, 
$p_i = \max \{  - \epsilon, - x_i \} $ and
$q_i = \min \{ \epsilon , 1 - x_i\}$ characterize the boundary of the linear constraint \eqref{eq: lin_cons} for the variable $\delta_i$,
and  $\mathbf e_i \in \mathbb R^d$ denotes the  basis vector  with the $i$th entry being $1$ and others being $0$.
\end{mycor}

\paragraph{{\FBAT} algorithm and implementation.}
Similar to {\FAT} or {\AT}, the {\FBAT} algorithm also follows the principle of alternating optimization. 
Specifically, it consists of 
the {\IG}-based 
upper-level gradient descent \eqref{eq: GD_upper},  interlaced with the lower-level optimal attack \eqref{eq: sol_linear}. 
We summarize the {\FBAT} algorithm below.

\noindent $\bullet$ \textit{Lower-level solution}: Obtain $\bdelta^*(\btheta_t)$ from  \eqref{eq: sol_linear};

{\vspace*{-5mm} \small 
\begin{align*}
\text{\resizebox{.3\textwidth}{!}{$
\bdelta^*(\btheta) = \mathcal P_{\mathcal C} \left(\mathbf z - (1/\lambda) \nabla_{\bdelta = \mathbf z }\ell_{\mathrm{atk}}( \btheta, \bdelta) \right).
$}}
\end{align*}}%
\vspace*{-7mm}

\noindent $\bullet$ \textit{Upper-level model training}: Integrating the {\IG} \eqref{eq: IG_fast_BAT} into \eqref{eq: GD_upper}, call SGD to update the model parameters as:

{\vspace*{-5mm}
\small
\begin{align}
\raisetag{15mm}
\text{\resizebox{.43\textwidth}{!}{
$\begin{array}{ll}
    \btheta_{t+1} = \btheta_t \hspace*{-3mm} & - \alpha_1 \nabla_{\btheta} \ell_{\mathrm{tr}}(\boldsymbol \theta_t, \boldsymbol \delta^\ast) \\
    & - \alpha_2 (-\frac{1}{\lambda})\nabla_{\btheta \bdelta} \ell_{\mathrm{atk}}(\btheta_t , \bdelta^\ast) \mathbf H_{\mathcal C}
    \nabla_{\bdelta} \ell_{\mathrm{tr}} (\boldsymbol \theta_t, \boldsymbol \delta^\ast) ,
    \hspace*{-3mm}   
\end{array}$
}}
\label{eq: SGD_fast_BAT}
\end{align}}%
\vspace*{-5mm}

\noindent where $\alpha_1, \alpha_2 > 0$ are learning rates associated with the model gradient and the {\IG}-augmented descent direction.

It is clear from \eqref{eq: SGD_fast_BAT} that to train a robust model, {\FBAT} can  be decomposed into   the regular {\FAT} update
(\textit{i.e.}, the term multiplied by $\alpha_1$)
and the additional update that involves {\IG}, (\textit{i.e.}, the term multiplied by $\alpha_2$). 
To implement {\FBAT}, we highlight some key hyper-parameter setups that are different from {\FAT} \citep{Wong2020Fast} and
{\FATGA} \citep{andriushchenko2020understanding}.

\paragraph{Remark on implementation details}
\label{remark2}
Next, we discuss the 
practical setups
used in Fast-BAT (more details in Appendix\,\ref{app: experiment_setting}).
{First, $1/\lambda$ serves as the attack step size as shown in \eqref{eq: sol_linear}. We refer readers to Table \ref{tab: lambda_sensitivity} for a   sensitivity analysis of $\lambda$.
In  Fast-BAT, the hyperparameter $\alpha_2$ is     newly introduced, and is set differently from $\alpha_1$ so as to control the   descent error associated with   the coupled second-order/first-order stochastic derivatives, \textit{i.e.}, the $\alpha_2$-term in \eqref{eq: SGD_fast_BAT}.}
In Sec.\,\ref{tab: lambda_sensitivity}, we show that a proper $\alpha_2$ helps mitigate catastrophic overfitting. %
In practice, we  set $\alpha_2/\lambda = 0.1  \alpha_1$.
To specify the linearization point $\mathbf z$ in \eqref{eq: prob_biLevel_lin},
we investigate two types of linearization schemes: (1) the random constant linearization (random uniform and random corner linearization) and (2) the $1$-step perturbation warm-up-based linearization ($1$-step sign-based and $1$-step  w/o sign PGD). 
These linearization schemes have  computational complexities up to the one-step attack generation.
Empirically, we find that {\FBAT} using ``$1$-step PGD w/o sign" leads to the best defensive performance (see Table\,\ref{tab: linearization_sensitivity}). 
We follow this experiment setup in the sequel. 

\paragraph{Remark on convergence analysis} 
In Appendix\,\ref{sec: convergence}, we prove that under some assumptions on the gradient bias, Fast-BAT converges to a first-order stationary point or its small neighborhood in the rate of $\mathcal{O} (1/\sqrt{T})$, where $T$ is the iteration number of model updates.
The main analysis difficulty lies in the
last term of the model updating rule   \eqref{eq: SGD_fast_BAT}, which involves two coupled derivatives built upon the same mini-batch.

\iffalse
\paragraph{\RV{Remark on the deficiency of {\FAT} compared to {\FBAT}.}} 
Signed gradient in {\FAT} seems effective to avoid IG (implicit gradient) in BLO. However, it only provides an {approximated} BLO  solution, since the discrete nature of the sign operation violates the `smoothness' assumption needed to apply chain rule in IG derivation. This necessitates the non-signed Taylor expansion and KKT conditions to derive IG (Proposition \ref{prop: 1}) rigorously. 
Besides, the MMO does not suffice to explain {\FAT} due to the gap between classical gradient descent-ascent optimization \cite{razaviyayn2020nonconvex} and the actual implementation of {\FAT}. BLO solves above issue as it can use a different lower-level objective than the upper level, leading to {new explanation} of {\FAT} (BLO + lower-level linearization).
\fi 

\section{Experiments}
\label{sec: exp}

\subsection{Experiment Setup}
\label{sec: exp_setup}

\paragraph{Datasets and model architectures.}
We will evaluate the effectiveness of our proposal under CIFAR-10~\citep{Krizhevsky2009learning}, CIFAR-100~\citep{Krizhevsky2009learning}, Tiny-ImageNet~\citep{deng2009imagenet}, and ImageNet~\citep{deng2009imagenet}. Unless  specified otherwise, we will train DNN models PreActResNet (PARN)-18~\citep{he2016identity}   for all datasets except ImageNet, and ResNet (RN)-50~\citep{he2016deep} for ImageNet. As a part of the ablation study, we also train larger models PARN-50 and WideResNet (WRN)-16-8~\citep{zagoruyko2016wide} on CIFAR-10.  Some preliminary ImageNet results are   reported in Appendix \ref{app: additional_result}.

\paragraph{Baselines.}
We focus on three baselines, namely, {\FAT}~\citep{Wong2020Fast}, {\FATGA}~\citep{andriushchenko2020understanding}, and {\ATF}~\citep{madry2018towards}, and refer readers to comparisons with more baselines ({\PGDS} \cite{madry2018towards}, {\bs} \cite{chen2020efficient}, {\free} \cite{shafahi2019adversarial}, {\ATTA} \cite{zheng2020efficient}, {\YOPO} \cite{zhang2019you}) in Appendix \ref{app: additional_result}.
Here {\ATF} and {\PGDS} stand for the $2$-step and $7$-step PGD attack-based {\AT}, respectively.
The primal criterion of baseline selection is computational complexity. All the methods except {\PGDS} consume the training time of the same order, while {\PGDS} serves as a reference to the performance of the non-accelerated robust training method. We stress that {\FATGA} is the strongest baseline to the best of our knowledge in terms of improving robustness and mitigating robust catastrophic overfitting \cite{andriushchenko2020understanding}. 
 
\paragraph{Training details.}
We choose the training perturbation strength $\epsilon\ \in \{8, 16\}/255$ for CIFAR-10, CIFAR-100, and Tiny-ImageNet; and $\epsilon = 2/255$  for ImageNet following \citep{Wong2020Fast,andriushchenko2020understanding}.
Throughout the experiments, we utilize an SGD optimizer with a momentum of $0.9$ and weight decay of $5 \times 10^{-4}$.
For CIFAR-10, CIFAR-100 and Tiny-ImageNet, we train each model for 20 epochs in total, where we use the cyclic scheduler to adjust the learning rate. The learning rate linearly ascends from $0$ to $0.2$ within the first $10$ epochs and then reduces to $0$ within the last $10$ epochs. Our batch size is set to 128 for all settings. 
In the implementation of {\FBAT}, 
we follow the dataset-agnostic hyperparameter scheme for $\lambda$, such that $\lambda = 255/5000$ for $\epsilon = 8/255$ and $\lambda = 255/2500$ for $\epsilon = 16/255$ for CIFAR-10, CIFAR-100 and Tiny-ImageNet. For ImageNet, we  strictly follow the setup given by \cite{Wong2020Fast} and we choose the train-time attack budget as $\epsilon = 2/255$.
For each method, we use the early stopping method to pick the model with the best {robust accuracy}, following \citep{rice2020overfitting}. All the experiments are conducted on a single  GeForce RTX 3090 GPU.All the baselines are trained with the recommended configurations in their official GitHub repos. We refer readers to Appendix\,\ref{app: experiment_setting} for more details on the training setup.

\paragraph{Evaluation details.}

For adversarial evaluation, we report \underline{r}obust test \underline{a}ccuracy (\textbf{RA}) of a learned model against PGD attacks \citep{madry2018towards} (\textbf{RA-PGD}). Unless otherwise specified, we set the test-time perturbation strength ($\epsilon$) the same as the train-time value, and take $50$-step PGD with $10$ restarts all the datasets. Since AutoAttack~\citep{croce2020reliable} is known as the strongest robust benchmark evaluation metric (given as an ensemble attack), we also measure robust accuracy against AutoAttack, termed \textbf{RA-AA}. Further, we measure the \underline{s}tandard \underline{a}ccuracy (\textbf{SA}) against natural  examples. Results are averaged over 10 independent trials. We would like to highlight that all the methods share the same batch size, epoch number, and training hardware. Thus, the time consumption per epoch reported in Table\,\ref{table: motivation_intro_overview} 
and Table\,\ref{tab: hessian_study} 
will serve as a fair indicator of the algorithm complexity of different methods.

\begin{table}[t]
\centering
% \vspace*{-2mm}
\caption{{SA, RA-PGD, and RA-AA of different robust training methods in the setup (CIFAR-10, PARN-18 training with $\epsilon = 8/255$) and (CIFAR-10, PARN-18 training with $\epsilon = 16/255$), respectively. All the results are averaged over 10 independent trials with different random seeds.}} \vspace*{1mm}
\label{tab: different_eval_epsilon}
\vspace*{0.1in}
\resizebox{0.48\textwidth}{!}{%
\begin{tabular}{cc>{\centering}p{1.8cm}>{\centering}p{1.8cm}>{\centering}p{1.8cm}>{\centering}p{1.8cm}>{\centering}p{1.8cm}}
\toprule[1pt]
\midrule
\multicolumn{6}{c}{{CIFAR-10, PARN-18 trained with $\epsilon = 8/255$}} \\ \midrule

\multicolumn{1}{c|}{\multirow{2}{*}{Method}} 
& \multicolumn{1}{c|}{\multirow{2}{*}{SA (\%)}} 
& \multicolumn{2}{c|}{RA-PGD (\%)}
& \multicolumn{2}{c}{RA-AA (\%)}\\
\multicolumn{1}{c|}{} 
& \multicolumn{1}{c|}{}                        
& \multicolumn{1}{c|}{$\epsilon=8$} 
& \multicolumn{1}{c|}{$\epsilon=16$}             
& \multicolumn{1}{c|}{$\epsilon=8$} 
& \multicolumn{1}{c}{$\epsilon=16$}
\\ \midrule

\multicolumn{1}{c|}{\FAT} 
& \multicolumn{1}{c|}{\textbf{82.39}\footnotesize{$\pm 0.44$}}
& 45.49 \footnotesize{$\pm 0.41$}
& \multicolumn{1}{c|}{9.56 \footnotesize{$\pm 0.26$}} 
& 41.87 \footnotesize{$\pm 0.15$}
& \multicolumn{1}{c}{7.91 \footnotesize{$\pm 0.06$}}
\\

\multicolumn{1}{c|}{\FATGA} 
& \multicolumn{1}{c|}{79.71\footnotesize{$\pm 0.44$}} 
& 47.27 \footnotesize{$\pm 0.42$}
& \multicolumn{1}{c|}{11.57\footnotesize{$\pm 0.32$}} 
& 43.24 \footnotesize{$\pm 0.27$}
& \multicolumn{1}{c}{9.48 \footnotesize{$\pm 0.15$}}
\\

\multicolumn{1}{c|}{\ATF} 
& \multicolumn{1}{c|}{81.97\footnotesize{$\pm 0.41$}} 
& 44.62 \footnotesize{$\pm 0.39$}
& \multicolumn{1}{c|}{9.39 \footnotesize{$\pm 0.32$}} 
& 41.73 \footnotesize{$\pm 0.20$}
& \multicolumn{1}{c}{7.54\footnotesize{$\pm 0.25$}}
\\

\rowcolor{Gray}
\multicolumn{1}{c|}{\FBAT} 
& \multicolumn{1}{c|}{79.97 \footnotesize{$\pm 0.12$}} 
& \textbf{48.83} \footnotesize{$\pm 0.17$}
& \multicolumn{1}{c|}{\textbf{14.00} \footnotesize{$\pm 0.21$}} 
& \textbf{45.19} \footnotesize{$\pm 0.12$}
& \multicolumn{1}{c}{\textbf{11.51} \footnotesize{$\pm 0.20$}}\\
\midrule

\multicolumn{6}{c}{{CIFAR-10, PARN-18 trained with $\epsilon = 16/255$}} \\ \midrule

\multicolumn{1}{c|}{\FAT} 
& \multicolumn{1}{c|}{44.15\footnotesize{$\pm 7.27$}}
& 37.17 \footnotesize{$\pm 0.74$}
& \multicolumn{1}{c|}{21.83 \footnotesize{$\pm $}1.32} 
& 31.66 \footnotesize{$\pm 0.27$}
& \multicolumn{1}{c}{12.49 \footnotesize{$\pm 0.33$}}
\\

\multicolumn{1}{c|}{\FATGA} 
& \multicolumn{1}{c|}{58.29 \footnotesize{$\pm 1.32$}} 
& 43.86 \footnotesize{$\pm 0.67$}
& \multicolumn{1}{c|}{26.01 \footnotesize{$\pm 0.16$}} 
& 38.69 \footnotesize{$\pm 0.56$}
& \multicolumn{1}{c}{17.97 \footnotesize{$\pm 0.33$}}
\\

\multicolumn{1}{c|}{\ATF} 
& \multicolumn{1}{c|}{68.04 \footnotesize{$\pm 0.30$}} 
& 48.79 \footnotesize{$\pm 0.31$}
& \multicolumn{1}{c|}{24.30 \footnotesize{$\pm 0.46$}} 
& 41.59 \footnotesize{$\pm 0.22$}
& \multicolumn{1}{c}{15.40 \footnotesize{$\pm 0.29$}}
\\

\rowcolor{Gray}
\multicolumn{1}{c|}{\FBAT} 
& \multicolumn{1}{c|}{\textbf{68.16} \footnotesize{$\pm 0.25$}} 
& \textbf{49.05} \footnotesize{$\pm 0.12$}
& \multicolumn{1}{c|}{\textbf{27.69} \footnotesize{$\pm 0.16$}} 
& \textbf{43.64} \footnotesize{$\pm 0.26$}
& \multicolumn{1}{c}{\textbf{18.79} \footnotesize{$\pm 0.24$}}
\\\midrule
\bottomrule[1pt]
\end{tabular}%
}
\end{table}

\begin{table}[t]
\newcolumntype{g}{>{\columncolor{Gray}}c}

\begin{center}
\caption{\footnotesize{Performance of different robust training methods under different model types. All the models are both trained and evaluated with the same perturbation strength $\epsilon$. 
}}
\label{tab: different_model}
\vspace*{0.1in}
\begin{threeparttable}
\resizebox{0.48\textwidth}{!}{
\begin{tabular}{c|c|cccc}
    \toprule[1pt]
    \midrule
    
    Model 
    & Method 
    & \begin{tabular}[c]{@{}c@{}} SA(\%) \\ ($\epsilon=8/255$) \end{tabular} 
    & \begin{tabular}[c]{@{}c@{}} RA-PGD(\%) \\ ($\epsilon=8/255$) \end{tabular} 
    & \begin{tabular}[c]{@{}c@{}} SA(\%) \\ ($\epsilon=16/255$) \end{tabular} 
    & \begin{tabular}[c]{@{}c@{}} RA-PGD(\%) \\ ($\epsilon=16/255$) \end{tabular} 
    \\ \midrule

    \multirow{4}{*}{PARN-50} 
    
    & {\FAT}     
    & 73.15\footnotesize{$\pm 6.10$}              
    & 41.03\footnotesize{$\pm 2.99$}                  
    & 43.86\footnotesize{$\pm 4.31$}                  
    & 22.08\footnotesize{$\pm 0.27$}  
    \\ 
    
    & {\FATGA}   
    & 77.40\footnotesize{$\pm 0.81$}                  
    & 46.16\footnotesize{$\pm 0.98$}                  
    & 42.28\footnotesize{$\pm 6.69$}           
    & 22.87\footnotesize{$\pm 1.25$} \\

    & {\ATF}     
    & \textbf{83.53}\footnotesize{$\pm 0.17$}                  
    & 46.17\footnotesize{$\pm 0.59$}                  
    & 68.88\footnotesize{$\pm 0.39$}           
    & 22.37\footnotesize{$\pm 0.41$}
    
    \\ 
    & \cellcolor{Gray} {\FBAT}
    & \cellcolor{Gray} 78.91\footnotesize{$\pm 0.68$}                  
    & \cellcolor{Gray} \textbf{49.18}\footnotesize{$\pm 0.35$}                  
    & \cellcolor{Gray} \textbf{69.01}\footnotesize{$\pm 0.19$}           
    & \cellcolor{Gray} \textbf{24.55}\footnotesize{$\pm 0.06$}
    
    \\
    \midrule
    
    \multirow{4}{*}{WRN-16-8} 
  
    & {\FAT}     
    & 84.39\footnotesize{$\pm 0.46$}                  
    & 45.80\footnotesize{$\pm 0.57$}                  
    & 49.39\footnotesize{$\pm 2.17$}           
    & 21.99\footnotesize{$\pm 0.41$} \\

    & {\FATGA}   
    & 81.51\footnotesize{$\pm 0.38$}                  
    & 48.29\footnotesize{$\pm 0.20$}                  
    & 45.95\footnotesize{$\pm 13.65$}          
    & 23.10\footnotesize{$\pm 3.90$} \\

    & {\ATF}     
    & \textbf{85.52}\footnotesize{$\pm 0.14$}                  
    & 45.47\footnotesize{$\pm 0.14$}                  
    & \textbf{72.11}\footnotesize{$\pm 0.33$}           
    & 23.61\footnotesize{$\pm 0.16$}

    \\
    & \mycolorrow {\FBAT}    
    & \mycolorrow 81.66\footnotesize{$\pm 0.54$}
    & \mycolorrow \textbf{49.93}\footnotesize{$\pm 0.36$}                  
    & \mycolorrow 68.12\footnotesize{$\pm 0.47$} 
    & \mycolorrow \textbf{25.63}\footnotesize{$\pm 0.44$}
    \\
    \midrule
    \bottomrule[1pt]
    \end{tabular}
}
\end{threeparttable}
\end{center}
\end{table}

\subsection{Results}
\label{sec: exp_robustness}

\paragraph{Overall performance of {\FBAT}.}

In Table~\ref{table: motivation_intro_overview}, Table\,\ref{tab: different_eval_epsilon}, and Table\,\ref{tab: imagenet}, 
we compare the overall performance of our proposed {\FBAT} with baselines.

\ding{172} We find that {\FBAT} consistently outperforms the other baselines  across the datasets and attack types.
In Table~\ref{table: motivation_intro_overview}, {\FBAT} improves the RA-PGD  performance consistently by over $1.5\%$ and RA-AA by around $1\%$ across all the datasets on both attack strengths. For stronger attacks with a larger perturbation budget, the advantage of {\FBAT} is even clearer, \textit{e.g.} a gain of over $2\%$ in Tiny-ImageNet with $\epsilon=16/255$.
On ImageNet, {\FBAT} outperforms {\FAT} by 1.23\% with $\epsilon = 2/255$.

\ding{173} {\FBAT} leads to a much better SA-RA trade-off compared with the baselines. For example, in Table~\ref{table: motivation_intro_overview}, the improvement in RA is not at cost of a huge drop in SA. Instead, when models are trained with $\epsilon = 16/255$, {\FBAT} even enjoys a significant boost over {\FATGA} in SA by $9.9\%$, $6.7\%$, and $4.6\%$ for CIFAR-10, CIFAR-100, and Tiny-ImageNet respectively.

\paragraph{Performance under different model architectures.} 
Besides PARN-18 reported above, Table~\ref{tab: different_model} presents results on both deeper (PARN-50) and wider (WRN-18-6) models.
As we can see, {\FBAT} consistently yields   {RA} improvement over other methods. We also note that  {\ATF} could be a   competitive baseline in terms of {SA}.
In contrast to {\FAT} and {\FATGA}, {\FBAT} is the only approach that yields an evident RA improvement over {\ATF}.

\begin{figure}[t]
\centerline{
\begin{tabular}{cc}
\hspace*{-2mm}\includegraphics[width=.23\textwidth,height=!]{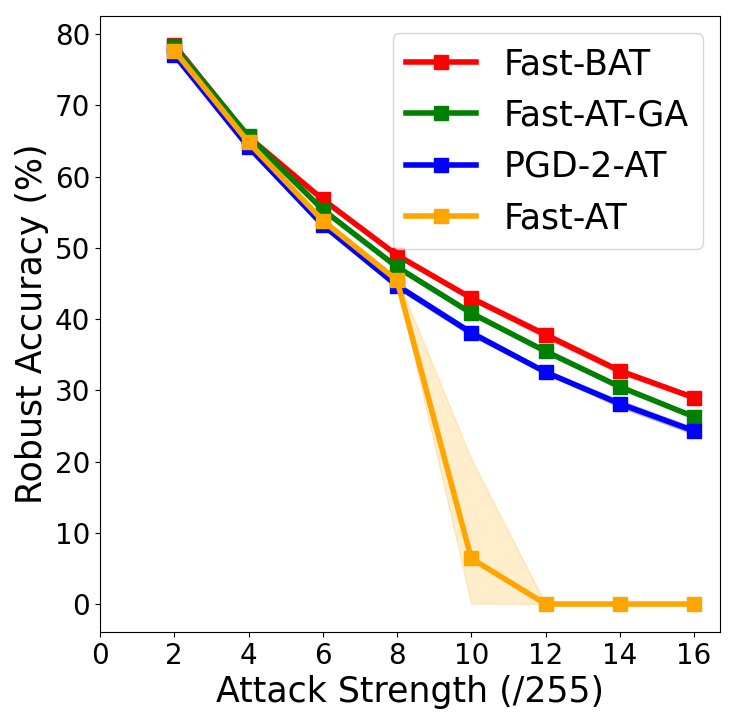}  
&\hspace*{-2mm}
\includegraphics[width=.23\textwidth,height=!]{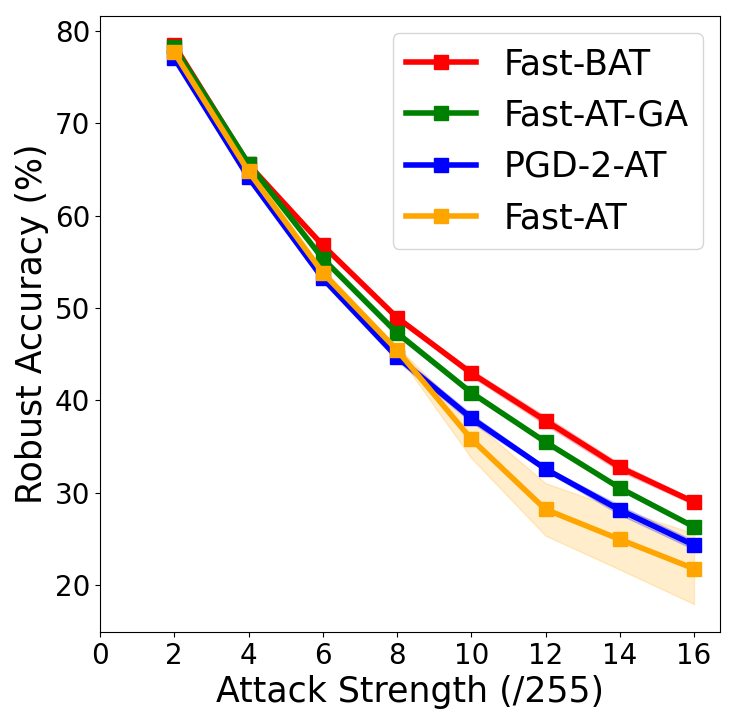} \vspace*{-2mm}\\
\footnotesize{(a) Without early stopping} &   \footnotesize{(b) With early stopping}
\end{tabular}}
\caption{\footnotesize{RA-PGD of different robust training methods for CIFAR-10 with the same training and evaluation attack strengths. 
}}\vspace*{-3mm}
\label{fig: early_stop}
\vspace*{-3mm}
\end{figure}

\paragraph{Mitigation of robust catastrophic overfitting.}
As shown in \citep{andriushchenko2020understanding}, {\FAT} suffers robust catastrophic overfitting when the train-time and test-time attack strength $\epsilon$ grows. 
Following \citep{andriushchenko2020understanding}, Figure\,\ref{fig: early_stop} presents two   RA-PGD trajectories, \textit{i.e.}, training without early stopping and training with early stopping, versus the train/test-time $\epsilon$.
As we can see, {\FAT} encounters a sharp RA drop when $\epsilon > 8$ when   early stopping is not used, consistent with the {\FATGA} paper \citep{andriushchenko2020understanding}. With early stopping, the overfitting of RA can be alleviated to some extent for  {\FAT}, but its performance still remains the worst. Moreover, different from \citep{andriushchenko2020understanding}, we find that {\ATF} yields resilient performance against catastrophic overfitting. Our implementation gives a more positive baseline for {\ATF}, since the implementation in \citep{andriushchenko2020understanding} did not use random initialization to generate train-time attacks.
Furthermore,    Figure\,\ref{fig: early_stop}  shows   that {\FBAT} mitigates the issue of catastrophic overfitting and yields improved RA over other methods. 
We highlight that such an achievement is \emph{free} of any robustness stability regularization, like gradient alignment   used in {\FATGA}.

\paragraph{Gradient alignment for `free'.}
As shown by \cite{andriushchenko2020understanding}, \emph{catastrophic overfitting} occurs with the local non-linearity of deep networks, which can be measured by the \underline{g}radient \underline{a}lignment (GA) score:

{\footnotesize
\vspace{-8mm}
\begin{align*}
    \mathbb E_{(\mathbf x, y)\sim \mathcal{D}, \boldsymbol \eta \sim \mathcal{U}({[-\epsilon, \epsilon]}^d)} [\text{cos}(\nabla_{\mathbf x} \ell_{\mathrm{tr}} (\mathbf x, y; \btheta), \nabla_{\mathbf x} \ell_{\mathrm{tr}} (\mathbf x + \boldsymbol \eta, y; \btheta))],
\end{align*}}%
\vspace*{-8mm}

where $\mathcal U$ denotes the randomly uniform distribution.
\begin{figure}
\vspace*{2mm}
\centerline{\hspace*{-4mm}
\includegraphics[width=0.3\textwidth,height=!]{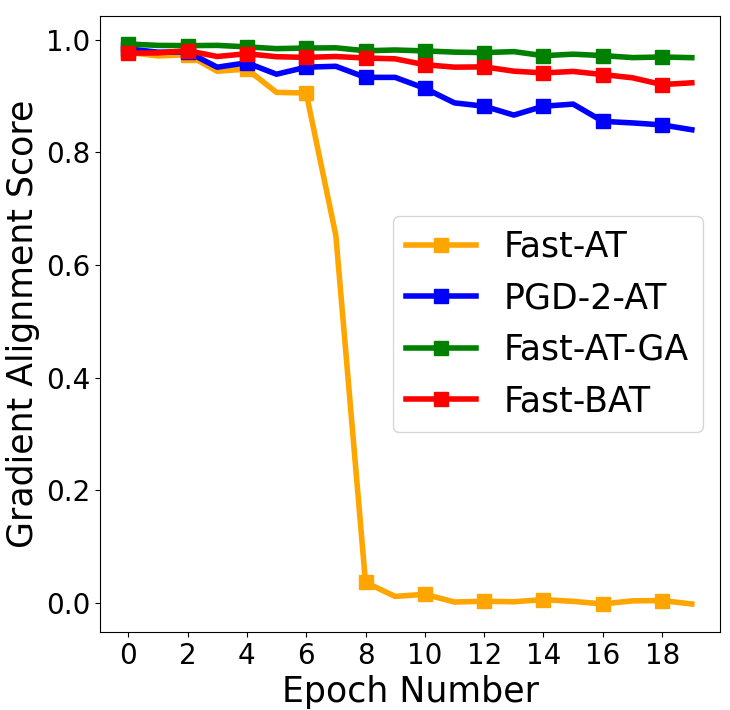}  
}
\vspace*{-2mm}
\caption{\footnotesize{
The training curve of GA score over different training methods on CIFAR-10.}}
\vspace*{-3mm}
\label{fig: ga_trajectory}
\end{figure}
GA is a key performance indicator to measure the appearance  of robustness catastrophic overfitting, as catastrophic overfitting is always accompanied by a sharp GA drop in the training trajectory of the robustly trained model.
In our paper, we calculate the GA for each method on the test set at the end of each epoch throughout the training process.
Figure\,\ref{fig: early_stop} suggested that {\FBAT} can mitigate overfitting without explicit GA regularization, and Figure~\ref{fig: ga_trajectory} presents the GA score versus the  training epoch number.
As can be seen, {\FBAT} automatically enforces GA and 
remains very close to {\FATGA}, which maximizes GA with an explicit regularization. Therefore,
a high GA score may just be a necessary but not a sufficient condition for avoiding catastrophic overfitting. 
As additional evidence, Fig.\,\ref{fig: loss_landscape} shows that similar to {\FATGA}, {\FBAT} has a flatter loss landscape than {\FAT} as well.
Therefore, a direct penalization on the input gradient norm may not achieve the state-of-the-art model robustness.

\begin{table}[t]
\begin{center}
\caption{\footnotesize{Robustness against adaptive attacks (RA-PGD) and transfer attacks (RA-Transfer Attack). 
Naturally trained PARN-18, PARN-50, and WRN-16-8 serve as source victim models for attack generation with PGD-20 ($\epsilon=8/255$) and PARN-18 robustified by different methods as target model for transfer attack evaluation.
}}
\label{tab:transfer_attack}
\vspace*{0.1in}
\begin{threeparttable}
\resizebox{0.47\textwidth}{!}{
\begin{tabular}{c|c|ccc}
\toprule[1pt]
\midrule
\multirow{2}{*}{Method} & \multirow{2}{*}{\begin{tabular}[c]{@{}c@{}}RA-PGD(\%)\end{tabular}} & \multicolumn{3}{c}{RA-Transfer Attack(\%)} \\
&   & PARN-18 & PARN-50 & WRN-16-8 \\ \midrule
{\FAT}                      
& 45.44\footnotesize{$\pm 0.06$}    
& 76.35\footnotesize{$\pm 0.12$}   
& 76.94\footnotesize{$\pm 0.14$}   
& 77.23\footnotesize{$\pm 0.21$} \\

{\ATF}                      
& 44.71\footnotesize{$\pm 0.04$}     
& 77.56\footnotesize{$\pm 0.14$}     
& 78.64\footnotesize{$\pm 0.12$}   
& 78.84\footnotesize{$\pm 0.17$} \\

{\FATGA}                    
& 47.31\footnotesize{$\pm 0.05$}    
& 77.34\footnotesize{$\pm 0.13$}   
& 78.34\footnotesize{$\pm 0.13$}   
& 78.53\footnotesize{$\pm 0.12$} \\ 

\rowcolor{Gray} \textbf{\FBAT (Ours)}     
& \textbf{48.67}\footnotesize{$\pm 0.05$}  
& \textbf{78.03}\footnotesize{$\pm 0.15$}   
& \textbf{79.93}\footnotesize{$\pm 0.12$}      
& \textbf{79.21}\footnotesize{$\pm 0.15$} \\ \midrule
\bottomrule[1pt]
\end{tabular}}
\end{threeparttable}
\end{center}
\end{table}

\paragraph{Sanity check for obfuscated gradients}
As pointed out by \cite{athalye2018obfuscated}, model robustness could be overestimated due to obfuscated gradients.  The model with obfuscated gradients  could   have `obfuscated' stronger resilience to white-box  (adaptive) attacks than  black-box (transfer) attacks. 
To justify the validity of {\FBAT}, Table\,\ref{tab:transfer_attack} summarizes the comparison between our proposal and the other baselines when facing adaptive and transfer attacks. Firstly, for all the methods,
RA increases if the transfer attack is applied, implying that the transfer attack is weaker than the adaptive attack. This is desired in the absence of obfuscated gradients. Moreover, {\FBAT} consistently outperforms the other baselines against both adaptive and transfer attacks. The absence of obfuscated gradients can also be justified by   
the flatness of the adversarial loss landscape in Figure.\,\ref{fig: loss_landscape}.

\paragraph{The validity of the Hessian-free assumption on ReLU-based neural networks.}

\begin{table}[t]
\vspace*{-2mm}
\centering
\caption{\footnotesize{Performance of Hessian-free and Hessian-aware {\FBAT} on CIFAR-10. We train and evaluate with the same attack budgets $\epsilon=8/255$ and $\epsilon=16/255$ to show the influence brought by Hessian matrix. 
}}
\label{tab: hessian_study}
\vspace*{0.1in}
\resizebox{.47\textwidth}{!}{%
\begin{tabular}{c|ccccccc}
\toprule[1pt]
\midrule
\textbf{Method} 
& \textbf{\begin{tabular}[c]{@{}c@{}}SA(\%)\\ ($\epsilon=8/255$)\end{tabular}} 
& \textbf{\begin{tabular}[c]{@{}c@{}}RA-PGD(\%)\\ ($\epsilon=8/255$)\end{tabular}} 
& \textbf{\begin{tabular}[c]{@{}c@{}}SA(\%)\\ ($\epsilon=16/255$)\end{tabular}} 
& \textbf{\begin{tabular}[c]{@{}c@{}}RA-PGD(\%)\\ $(\epsilon=16/255)$\end{tabular}} 
& \textbf{\begin{tabular}[c]{@{}c@{}}Time\\ (s/epoch)\end{tabular}} \\ \midrule

Hessian-free
& 79.97 \footnotesize{$\pm 0.12$}
& 48.83 \footnotesize{$\pm 0.17$}
& 68.16 \footnotesize{$\pm 0.25$}
& 27.69 \footnotesize{$\pm 0.16$}
& 61.4 \\
Hessian-aware
& 79.62 \footnotesize{$\pm 0.17$}
& 49.13 \footnotesize{$\pm 0.14$}
& 67.82 \footnotesize{$\pm 0.23$}
& 27.82 \footnotesize{$\pm 0.19$}
& 82.6 \\ \midrule
\bottomrule[1pt]
\end{tabular}%
}
\vspace*{-5mm}
\end{table}

In Corollary\,\ref{eq: thr_IG}, the Hessian-free assumption, \textit{i.e.}$\nabla_{\bdelta \bdelta}\ell_{\mathrm{atk}} = 0$, was made to simplify the computation of implicit gradient. 
To justify this assumption
we conduct experiments to compare the Hessian-free {\FBAT} with the Hessian-aware version. In Hessian-aware \FBAT, the implicit gradient is calculated based on~\eqref{eq: IG_cons}. In Table~\ref{tab: hessian_study}, the results do not indicate much difference when Hessian is used. However, the extra computations required to evaluate the Hessian heavily slows down {\FBAT} as around $30\%$ more time is needed. Therefore, the Hessian-free assumption is reasonable and also necessary in terms of the efficiency of the algorithm. 
We also justify this assumption on some non-ReLU neural networks and for more results please refer to Appendix\,\ref{app: additional_result}.

\section{Conclusion}
In this paper, we introduce a novel bi-level optimization-based fast adversarial training framework, termed {\FBAT}.  
The rationale behind designing {\FBAT}
lies in two aspects. 
First, from the perspective of implicit gradients, we show that the existing {\FAT} framework is equivalent to the lower-level linearized BLO along the sign direction of the input gradient.
Second, we show that {\FBAT} 
is able to achieve improved stability of performance, mitigated catastrophic overfitting, and enhanced accuracy-robustness trade-off. To the best of our knowledge, we for the first time establish the theory and the algorithmic foundation of BLO for adversarial training.
Extensive experiments are provided to demonstrate the superiority of our method to state-of-the-art accelerated {\AT} baselines.

\section*{\RV{Acknowledgement}}
Y. Zhang and S. Liu are supported by the Cisco Research   grant CG\# 70614511. M. Hong and P. Khanduri are supported in part by NSF grants CIF-1910385 and NSF CMMI-1727757.

\bibliography{refs, reference, ref_SL_adv, ref_SL_fair_self, ref_SL_ZO, ref_SL_AVs, ref-bi, ref_SL_pruning}
\bibliographystyle{icml2022}

\appendix
\onecolumn
%%%%%%%%%%%%%%%%%%%%%%%%%%%Appendix%%%%%%%%%%%%%%%%%%%%%%%%%%%%%%%%%
\newpage
\clearpage

\makeatletter 
\makeatother

\section{Proof of Proposition\,\ref{prop: 1} and Corollary\,\ref{eq: thr_IG}}
\label{app: IG_thr}

\textbf{Proof}:
Upon defining $g(\btheta, \bdelta) = (\bdelta - \mathbf z)^\top  \nabla_{\bdelta = \mathbf z}\ell_{\mathrm{atk}}( \btheta, \bdelta)   + (\lambda/2) \| \boldsymbol \delta -\mathbf z \|_2^2$, we   repeat  \eqref{eq: prob_biLevel_lin} as
 			\begin{align}
\begin{array}{ll}
\displaystyle \minimize_{\boldsymbol \theta }         & \mathbb E_{(\mathbf x, y) \in \mathcal D} [ \ell_{\mathrm{tr}}(\boldsymbol \theta,  \boldsymbol \delta^*(\boldsymbol \theta)) ] \\
\st  & \boldsymbol \delta^*(\boldsymbol \theta) = 
\displaystyle \argmin_{ \mathbf B \bdelta \leq \mathbf b }  g(\btheta, \bdelta),
\end{array}
\label{eq: prob_biLevel_lin_app}
%\tag{\text{Fast BAT}}
\end{align} 
where we have used the expression of linear constraints in \eqref{eq: lin_cons}.  

Our goal is to derive the {\IG} 
$\frac{d \bdelta^*(\btheta)^\top}{d \btheta}$ shown in \eqref{eq: GD_upper}. 
To this end,  we first build implicit functions by leveraging   KKT conditions of the lower-level   problem  
of \eqref{eq: prob_biLevel_lin_app}.
We say $\bdelta^*(\btheta)$ and $\boldsymbol \lambda^*(\btheta)$ (Lagrangian multipliers) satisfy the KKT conditions:
\begin{align}
    \label{eq: KKT}
    \begin{split}
   \text{Stationarity:} \qquad & \nabla_{\bdelta} g(\btheta , \bdelta^*(\btheta) ) + \mathbf  B^\top  \boldsymbol \lambda^\ast( \btheta ) = \mathbf 0, \\
    \text{Complementary slackness %(\textcolor{cyan}{not precise?})
    :
    } 
    \qquad & \boldsymbol \lambda^\ast(\btheta) \cdot (\mathbf B  \bdelta^\ast(\btheta) - \mathbf b) = \mathbf 0 \\
    \text{Dual feasibility:} \qquad & \boldsymbol \lambda^\ast(\btheta) \geq \mathbf  0
      \end{split}
\end{align}
where $\cdot$ denotes the elementwise product. 
\vspace{2 mm}

\noindent
\textit{Active constraints \& definition of $\mathbf B_0$}:
Let $\mathbf B_0$ denote the sub-matrix of $\mathbf B$ and $\mathbf b_0$ the sub-vector of $\mathbf b$, which consists of only the \textit{active constraints} at $\bdelta^\ast(\btheta)$, \textit{i.e.}, those satisfied with the equality $\mathbf B_0 \bdelta^\ast(\btheta) = \mathbf b_0$ (corresponding to \textit{nonzero} dual variables). The determination of active constraints is done given $\btheta$ %\textcolor{cyan}{$\bdelta_k$ you mean?} 
at each iteration.

With the aid of $(\mathbf B_0, \mathbf b_0)$, KKT \eqref{eq: KKT} becomes
\begin{align}
    \label{eq: KKT_v2}
    \nabla_{\bdelta} g(\btheta , \bdelta^\ast(\btheta)) + \mathbf  B_0^\top \boldsymbol \lambda^\ast( \btheta) = \mathbf 0, ~~  \text{and} ~~\mathbf B_0 \bdelta^\ast(\btheta) - \mathbf b_0 = \mathbf 0,
\end{align}
where the nonzero $\boldsymbol \lambda^\ast( \btheta)$ only corresponds to the active constraints. 
We take derivatives w.r.t. $\btheta$ of \eqref{eq: KKT_v2}, and thus obtain the following
\begin{align}
 &   \frac{d \nabla_{\bdelta} g(\btheta , \bdelta^\ast(\btheta))^\top}{d \btheta}
   +  \nabla_{\btheta} \boldsymbol \lambda^\ast( \btheta)^\top\mathbf  B_0 = \mathbf 0 \nonumber \\
   \Longrightarrow & \nabla_{\btheta \bdelta} g(\btheta , \bdelta^\ast(\btheta)) +  \underbrace{\frac{d \bdelta^*(\btheta)^\top}{d \btheta}}_\text{IG} \nabla_{\bdelta \bdelta} g(\btheta , \bdelta^\ast(\btheta)) +  \nabla_{\btheta} \boldsymbol \lambda^\ast( \btheta)^\top\mathbf  B_0 = \mathbf 0,
  \label{eq: KKT1}  \\ 
\text{and }  & ~  \underbrace{\frac{d \bdelta^*(\btheta)^\top}{d \btheta}}_\text{IG} \mathbf B_0^\top= \mathbf 0,  \label{eq: KKT2} 
  %\mathbf B_0 \nabla_{\btheta}\bdelta^\ast(\btheta)   = \mathbf 0
 \end{align}
 where $\nabla_{\btheta \bdelta} \in \mathbb R^{|\btheta| \times |\bdelta|}$ denotes second-order partial derivatives (recall that $|\btheta| = n$ and $ |\bdelta| = d$).  
 According to \eqref{eq: KKT1}, we have
\begin{align}
    {\frac{d \bdelta^*(\btheta)^\top}{d \btheta}}  = - [\nabla_{\btheta \bdelta} g(\btheta , \bdelta^\ast(\btheta))  +\nabla_{\btheta} \boldsymbol \lambda^\ast( \btheta)^\top\mathbf  B_0 ] \nabla_{\bdelta \bdelta} g(\btheta , \bdelta^\ast(\btheta))^{-1}.
    \label{eq: KKT3}
\end{align}
Substituting the above into \eqref{eq: KKT2}, we obtain 
\begin{align}
     \nabla_{\btheta \bdelta} g(\btheta , \bdelta^\ast(\btheta))\nabla_{\bdelta \bdelta} g(\btheta , \bdelta^\ast(\btheta))^{-1} \mathbf B_0^\top +\nabla_{\btheta} \boldsymbol \lambda^\ast( \btheta)^\top\mathbf  B_0 \nabla_{\bdelta \bdelta} g(\btheta , \bdelta^\ast(\btheta))^{-1} \mathbf B_0^\top = \mathbf 0,
\end{align}
which yields:
\begin{align}
    \nabla_{\btheta} \boldsymbol \lambda^\ast( \btheta)^\top  = -  \nabla_{\btheta \bdelta} g(\btheta , \bdelta^\ast(\btheta)) \nabla_{\bdelta \bdelta} g(\btheta , \bdelta^\ast(\btheta))^{-1} \mathbf{B}_0^\top[\mathbf B_0 \nabla_{\bdelta \bdelta} g(\btheta , \bdelta^\ast(\btheta))^{-1} \mathbf B_0^\top]^{-1} ,
    \label{eq: KKT4}
\end{align}
and thus,
\begin{align}
    \nabla_{\btheta} \boldsymbol \lambda^\ast( \btheta)^\top\mathbf B_0 = -  \nabla_{\btheta \bdelta} g(\btheta , \bdelta^\ast(\btheta)) \nabla_{\bdelta \bdelta} g(\btheta , \bdelta^\ast(\btheta))^{-1} \mathbf{B}_0^\top[\mathbf B_0 \nabla_{\bdelta \bdelta} g(\btheta , \bdelta^\ast(\btheta))^{-1} \mathbf B_0^\top]^{-1} \mathbf B_0.
    \label{eq: KKT5}
\end{align}

Substituting \eqref{eq: KKT5} into \eqref{eq: KKT3}, {we obtain the IG}
\begin{align}
 &  {\frac{d \bdelta^*(\btheta)^\top}{d \btheta}}
   =  -\nabla_{\btheta \bdelta} g(\btheta , \bdelta^\ast(\btheta)) \nabla_{\bdelta \bdelta} g(\btheta , \bdelta^\ast(\btheta))^{-1} -  \nabla_{\btheta} \boldsymbol \lambda^\ast( \btheta)^\top \mathbf B_0\nabla_{\bdelta \bdelta} g(\btheta , \bdelta^\ast(\btheta))^{-1} \nonumber \\
&    =  -\nabla_{\btheta \bdelta} g(\btheta , \bdelta^\ast(\btheta)) \nabla_{\bdelta \bdelta} g(\btheta , \bdelta^\ast(\btheta))^{-1} \nonumber \\
   & + \nabla_{\btheta \bdelta} g(\btheta , \bdelta^\ast(\btheta)) \nabla_{\bdelta \bdelta} g(\btheta , \bdelta^\ast(\btheta))^{-1} \mathbf{B}_0^\top [\mathbf B_0 \nabla_{\bdelta \bdelta} g(\btheta , \bdelta^\ast(\btheta))^{-1} \mathbf B_0^\top]^{-1} \mathbf B_0 \nabla_{\bdelta \bdelta} g(\btheta , \bdelta^\ast(\btheta))^{-1}.
  \label{eq: IG_cons}
\end{align}

The above completes the proof of Proposition\,\ref{prop: 1}. \hfill $\square$

To further compute \eqref{eq: IG_cons}, the Hessian matrix $\nabla_{\bdelta \bdelta} \ell_{\mathrm{atk}}$ is needed. Recall from  the definition of the lower-level objective   that the Hessian matrix is   given by
\begin{align}\label{eq: Hessian_simplification}
    \nabla_{\bdelta \bdelta} g(\btheta , \bdelta^\ast(\btheta)) = \nabla_{\bdelta \bdelta} \ell_{\mathrm{atk}} +\lambda \mathbf I = \mathbf 0 + \lambda \mathbf I.
\end{align}
Here we used the assumption that 
$\nabla_{\bdelta \bdelta} \ell_{\mathrm{atk}} = \mathbf 0$. The rationale behind that   is
     % ReLU-based
      neural networks commonly
      %(used in experiments) 
   leads to a piece-wise linear decision boundary w.r.t.
%neural network as a function of 
the inputs \citep{moosavi2019robustness,alfarra2020decision}, and thus, 
its   second-order derivative (Hessian) $\nabla_{\bdelta \bdelta}  \ell_{\mathrm{atk}}$  is close to zero.

 Based on the simplification \eqref{eq: Hessian_simplification}, we have 
\begin{align}
   {\frac{d \bdelta^*(\btheta)^\top}{d \btheta}} 
    = & - (1/\lambda)\nabla_{\btheta \bdelta} g(\btheta , \bdelta^\ast(\btheta)) \underbrace{ \left (    \mathbf I -    \mathbf{B}_0^\top [  \mathbf B_0     \mathbf B_0^\top]^{-1} \mathbf B_0    \right ) }_{\Def \mathbf H_{\mathcal C}} \nonumber \\
    &  - (1/\lambda)\nabla_{\btheta \bdelta} \ell_{\mathrm{atk}}(\btheta , \bdelta^\ast(\btheta)) \mathbf H_{\mathcal C},
    %\tag{IG-v1} 
    \label{eq: IG_cons_simplify}
\end{align}
where we have used the fact that $\nabla_{\btheta \bdelta} {{g}}   = \nabla_{\btheta \bdelta} \ell_{\mathrm{atk}}  $.
%where $\mathbf H$ may change over iterations according to $\mathbf B_0$ (\textit{i.e.}, active constraints).

\underline{What is $\mathbf H_{\mathcal C}$ in \eqref{eq: IG_cons_simplify}?}
Since $ \mathbf B = \begin{bmatrix}
    \mathbf I \\
     - \mathbf I
    \end{bmatrix}$, we can obtain that   $\mathbf B_0 \mathbf B_0^\top = \mathbf I$ and $\mathbf B_0^\top  \mathbf B_0$ is a sparse diagonal matrix with diagonal entries being $0$ or $1$. Thus,
    $\mathbf H_{\mathcal C}$ can be first simplified to
\begin{align}
    \mathbf H_{\mathcal C} = \mathbf I - \mathbf B_0^\top  \mathbf B_0.
\end{align}
Clearly, $\mathbf H_{\mathcal C}$ is also a diagonal matrix with either $0$ or $1$ diagonal entries. The $1$-valued diagonal entry of $\mathbf H_{\mathcal C}$ corresponds to the \textit{inactive constraints} in
$
\mathbf B  \bdelta^*(\btheta) < \mathbf b
$, \textit{i.e.}, those satisfied with \textit{strict inequalities} in $\{ \| \bdelta \|_\infty  \leq \epsilon , \mathbf 0 \leq \bdelta \leq \mathbf 1 \}$. This 
can be expressed as 
\begin{align}
\mathbf H_{\mathcal C} = \begin{bmatrix}
1_{p_1 \leq \delta_1^* \leq q_1  } \mathbf e_1, \ldots, 1_{p_1 \leq \delta_d^* \leq q_d  } \mathbf e_d
\end{bmatrix}
%  \max \{  - \epsilon, - x_1 \} \leq  z_1 \leq \min \{ \epsilon , 1 - x_1\}
    % \mathcal{P}_{\| \bdelta \|_{\infty} \leq \epsilon}(\mathbf z) = \max ( \min(\mathbf z, \epsilon), -\epsilon ),
    % ~~~ \text{yielding} ~~\frac{ d \mathcal P_{\mathcal C} (\mathbf z)  }{\partial z_i} = \left \{ 
    % \begin{array}{ll}
    %     \mathbf e_i & \text{if $| z_i | < \epsilon$} \\
    %   \mathbf 0  &  \text{otherwise},
    % \end{array}
    % \right.
    \label{eq: proj_ell_inf}
\end{align}
where 
$1_{p_i \leq \delta_i^* \leq q_i} \in \{ 0, 1\}$ denotes the indicator function over the constraint $\{ p_i \leq \delta_i^* \leq q_i \}$ and returns $1$ if the constraint is satisfied, $\delta_i^*$ denotes the $i$th entry of $\bdelta^*(\btheta)$, 
$p_i = \max \{  - \epsilon, - x_i \} $ and
$q_i = \min \{ \epsilon , 1 - x_i\}$, and  $\mathbf e_i \in \mathbb R^d$ denotes the  basis vector  with the $i$th entry being $1$ and others being $0$s.

Based on the definition of $g$, \eqref{eq: IG_cons_simplify} and \eqref{eq: proj_ell_inf}, we can eventually achieve the desired {\IG} formula \eqref{eq: IG_fast_BAT}.
The proof of Corollary\,\ref{eq: thr_IG} is now complete. 
\hfill $\square$

\newpage

%{\red[this section needs to be re-write; the problem is equivalent, but the gradient computation may not be, as stated in the main paper.]}

\section{Discussion on case $\ell_{\mathrm{atk}}=-\ell_{\mathrm{tr}}$}
\label{app: proof_problem2_problem1}

We provide an in-depth explanation on the fact that even if we set $\ell_\text{atk} = - \ell_\text{tr}$, the \textit{optimization routine} given by \eqref{eq: GD_upper}  to solve problem\,\eqref{eq: prob_biLevel} does not reduce to the ordinary IG-absent gradient descent to solve
problem\,\eqref{eq: prob_AT} because of the presence of lower-level constraints.

\begin{itemize}
    \item In the \textbf{absence} of the constraint $\boldsymbol \delta \in \mathcal C$, if we set $\ell_{\text{atk}} = -\ell_{\text{tr}}$, then solving problem\,\eqref{eq: prob_biLevel} via 
    IG-involved   descent method \eqref{eq: GD_upper}
    will reduce to the ordinary IG-absent method that solves problem\,\eqref{eq: prob_AT}. 
    
    This is a known BLO result (\emph{e.g.} \cite{ghadimi2018approximation}) and can be readily proven using the stationary condition. To be specific, based on the stationary condition of unconstrained lower-level optimization, we have $\nabla_{\boldsymbol \delta}\ell_{\text{atk}}(\boldsymbol\theta, \boldsymbol\delta^*) = 0$. Since $\ell_{\text{atk}} = -\ell_{\text{tr}}$, we have $\nabla_{\boldsymbol \delta}\ell_{\text{tr}}(\boldsymbol\theta, \boldsymbol\delta^*) = 0$. As a result, the second term in \eqref{eq: GD_upper} becomes $\mathbf 0$ and solving problem\,\eqref{eq: prob_biLevel} becomes identical to solving the min-max problem\,\eqref{eq: prob_AT}. 
    \item In the \textbf{presence} of the constraint $\boldsymbol \delta \in \mathcal C$, the stationary condition cannot be applied since the stationary point may not be a feasible point in the constraint. In other words, $\nabla_{\boldsymbol \delta}\ell_{\text{atk}}(\boldsymbol\theta, \boldsymbol\delta^*) = 0$ does not hold in the case of $\ell_{\text{atk}} = -\ell_{\text{tr}}$. As a matter of fact, one has to resort to KKT conditions instead of the stationary condition for a constrained lower-level problem. Similar to our proof in Proposition\,\ref{prop: 1}, the implicit gradient (and thus the second term of \eqref{eq: GD_upper}) cannot be omitted in general. This makes the optimization routine to solve problem\,\eqref{eq: prob_biLevel} different from solving   problem\,\eqref{eq: prob_AT}.
\end{itemize}

% \section{\RV{Derivation of Implicit Gradient for \FAT}}
% \label{app: der_IG_Fast_AT}
% We can derive \eqref{eq: IG_Fast_AT} using KKT condition similar to Theorem\,\ref{eq: thr_IG}. Specifically, let 
% \begin{align}
% g(\btheta, \bdelta) = \inp{\mathrm{sign}(\nabla_{\bdelta = \mathbf z}\ell_{\mathrm{atk}}( \btheta, \bdelta; \mathbf x, y) )}{\bdelta - \mathbf z} + \frac{\lambda }{2} \| \boldsymbol \delta -\mathbf z \|_2^2,
% \label{eq: g_Fast_AT}
% \end{align}
% we have
% %\begin{align}
%  $  \nabla_{\btheta \bdelta} g  = \mathbf 0 $ almost surely{\red[why? can you elaborate?]}.
%   % \label{eq: so_derivative_g_Fast_AT}
% %\end{align}
% Following \eqref{eq: IG_cons}, we can further obtain \eqref{eq: IG_Fast_AT}.

% {\red[I suggest to write down how this is derived explicitly, and say what is the approximation that we are using regarding the gradient and Hessian of the sign operator. When reader read the paper, they will come here to look for proofs, and there is none. If you refer them to the proof in Theorem 1, how are they going to read? Theorem 1 has not been discussed so far. Please note that our computation is not precise, it critically depends on these approximations.]}

\newpage

\section{Detailed Experiment settings}
\label{app: experiment_setting}
\subsection{Training Set-up}

% \SL{[Put the following in appendix, called Training setup]}
For CIFAR-10, we summarize the training setup for each method. 1) \underline{\FAT}: We use FGSM with an attack step size of $1.25 \epsilon$ to generate perturbations;
2) \underline{\ATF}: 2-step PGD attacks\footnote{We use random initialization to generate perturbations for PGD, while in the paper of \FATGA~\citep{andriushchenko2020understanding}, 2-step PGD is initialized at zero point, which we believe will underestimate  the effect of \ATF} with an attack step size of $0.5 \epsilon$ is implemented;
3) \underline{\FATGA}: The gradient alignment regularization parameter is set to the recommended value for each $\epsilon$;
4) \underline{\FBAT}: We select $\lambda$ from $255/5000$ to $255/2000$ for different $\epsilon$.
At the same time, we adjust $\alpha_2$ accordingly, so that the coefficient of the second term in \eqref{eq: SGD_fast_BAT}, namely $\alpha_2 /\lambda$ always equals to $0.1 \alpha_1$. 

For ImageNet, we set $\epsilon$ to $2/255$ , and we strictly follow the training setting adopted by \cite{Wong2020Fast}.
% In {\FBAT}, we fix $\lambda$ at $255/3000$ and adopt the same $\alpha_2$ selection strategy as CIFAR-10. 
In {\FBAT}, we fix $\lambda$ at $255/3000$ and adopt the same $\alpha_2$ selection strategy as CIFAR-10. 
%Detailed experimental settings in ImageNet are specified in Appendix\, \ref{app: experiment_setting}.

% We refer readers to \YH{Appendix\,\ref{app:  experiment_setting}} for more details on hyper-parameter selection.

\paragraph{Parameter for {\FATGA}} Regarding {\FATGA} with different model types, we adopt the same regularization parameter recommended in its official repo\footnote{\RV{\FATGA: \url{https://github.com/tml-epfl/understanding-fast-adv-training/blob/master/sh}}} intended for PreActResNet-18 (namely 0.2 for $\epsilon = 8/255$ and 2.0 for $\epsilon = 16/255$). 
% \SL{Move this to appendix, combine it with implementation details.}

\subsection{The choice of initialization point $\mathbf{z}$}\label{sec: z_choice}
To specify $\mathbf z$ in  \eqref{eq: prob_biLevel_lin},
we investigate two classes of linearization schemes.
The first class is random constant linearization, including: ``uniformly random linearization'', i.e., $\mathbf z = \bdelta_0$ similar to {\FAT}, and  ``random corner linearization" under the $\epsilon$-radius $\ell_\infty$-ball, i.e., $\mathbf z \in \{ -\epsilon, \epsilon  \}^d$. 
The second class is $1$-step perturbation warm-up-based linearization, including the other two specifications:
``$1$-step sign-based PGD"
% $ \mathbf z = P_{\mathcal C} \left   ( \bdelta_0 + \alpha \cdot  \mathrm{sign} \left ( \nabla_{\bdelta} \ell_{\mathrm{tr}}(\btheta_t, \bdelta_0) \right  ) \right  )$
, and ``$1$-step PGD w/o sign"
% $ \mathbf z = P_{\mathcal C} \left   ( \bdelta_0 + \alpha  \nabla_{\bdelta} \ell_{\mathrm{tr}}(\btheta_t, \bdelta_0)  \right  )$
. 
We consider this linearization schemes as their computation complexities are less than or close to the complexity of one-step attack generation. As a result, {\FBAT}
% combined with these linearizations 
takes comparable computation cost to the baselines  {\FAT}, {\ATF} and {\FATGA}.
Empirically, we find that {\FBAT} using ``$1$-step PGD w/o sign" leads to the best defensive performance; see justification in Table\,\ref{tab: linearization_sensitivity}. We follow this experiment setup in the sequel.

\newpage
\section{Additional Experimental Results}
\label{app: additional_result}

% \paragraph{Preliminary ImageNet results}

% \begin{wraptable}{r}{55mm}
%     \vspace{-6mm}
%     \caption{\footnotesize{SA and RA on ImageNet.
%     % The results of {\FAT} are directly from the Table 4 in \cite{Wong2020Fast}. 
%     % {\FATGA} does not show benefits in ImageNet and its baseline in Table~6 in \cite{andriushchenko2020understanding} is lower than ours as shown. As they do not release their code for ImageNet, we do not make a direct comparison with {\FATGA} here.
%     }}
%     \label{tab: imagenet}
%     \centering
%     \vspace{1mm}
%     \resizebox{0.3\textwidth}{!}{
%     \begin{tabular}{c|c|c}
%     \toprule[1pt]
%     \midrule
%          Method &  SA (\%) & RA-PGD (\%) \\
%          \midrule
%          {\FAT} & 60.90 
%         %  \footnotesize{$\pm 0.47$} 
%          & 43.43 
%         %  \footnotesize{$\pm 0.24$} 
%          \\
%         %  {\ATF} &   60.89 & \textcolor{red}{\textbf{44.74}}\\
%          {\FBAT} &  60.18
%         %  \footnotesize{$\pm 0.09$} 
%          & 44.64 
%         %  \footnotesize{$\pm 0.11$} 
%          \\
%         %  \midrule
%         %  {\FAT} & $4/255$ & 55.45 & 30.18 \\
%         %  {\FBAT} & $4/255$ &  &  \\
%     \midrule
%     \bottomrule[1pt]
%     \end{tabular}}
%     \vspace{-3mm}
% \end{wraptable}

\paragraph{Results on ImageNet}

\begin{figure*}[htb]
\centering
\begin{adjustbox}{max width=\textwidth}
\begin{tabular}{@{\hskip 0.00in}c  @{\hskip 0.00in} @{\hskip 0.02in} c @{\hskip 0.02in}  @{\hskip 0.02in} c @{\hskip 0.02in}  @{\hskip 0.02in} 
}
 { \textbf{\footnotesize {\FAT} }}
& { \textbf{\footnotesize {\FATGA} }}
& { \textbf{\footnotesize {\FBAT}}} \\
\begin{tabular}{@{\hskip 0.02in}c@{\hskip 0.02in} }
\parbox[c]{10em}{\includegraphics[width=10em]{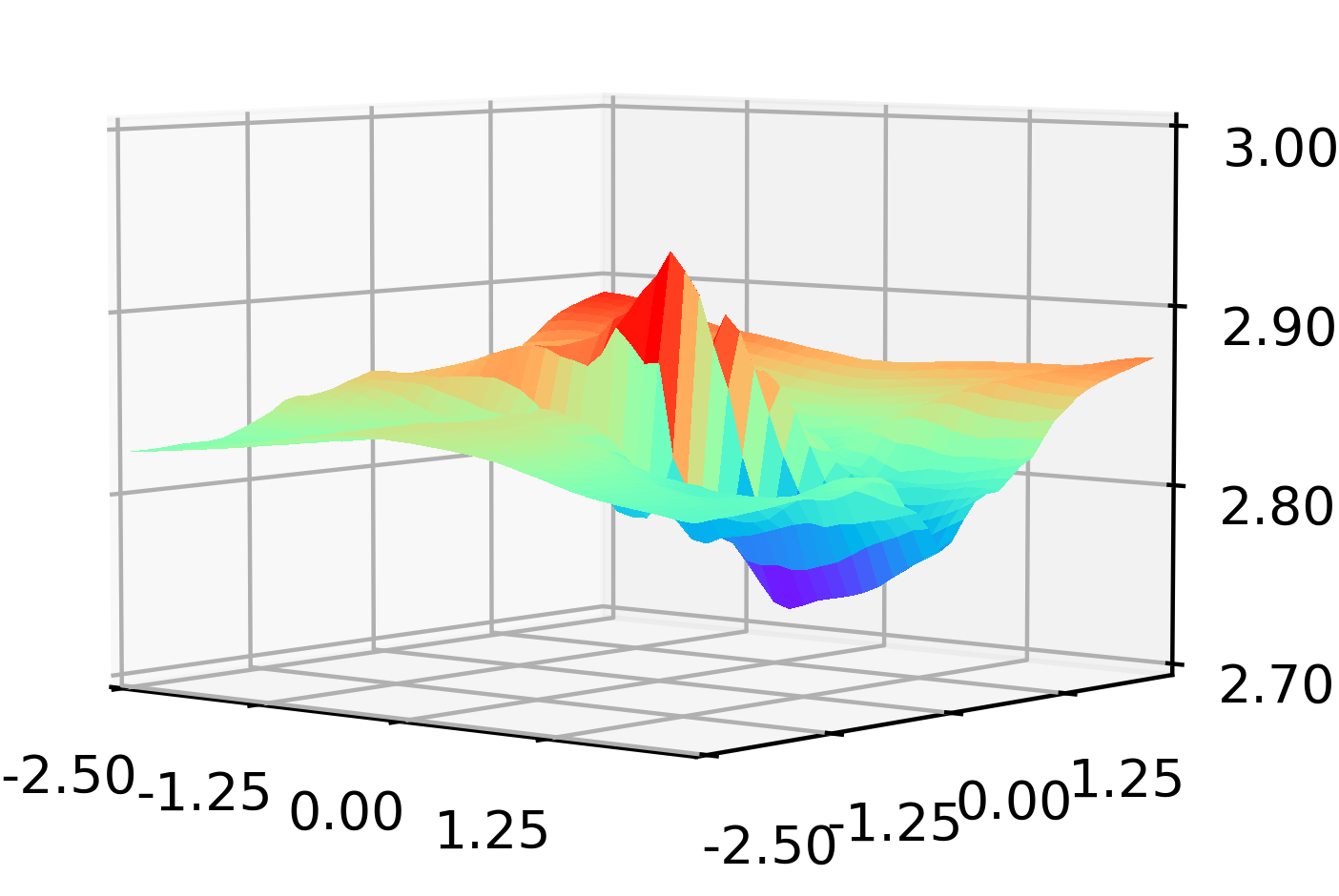}}  
\end{tabular}
& 
\begin{tabular}{@{\hskip 0.02in}c@{\hskip 0.02in} }
\parbox[c]{10em}{\includegraphics[width=10em]{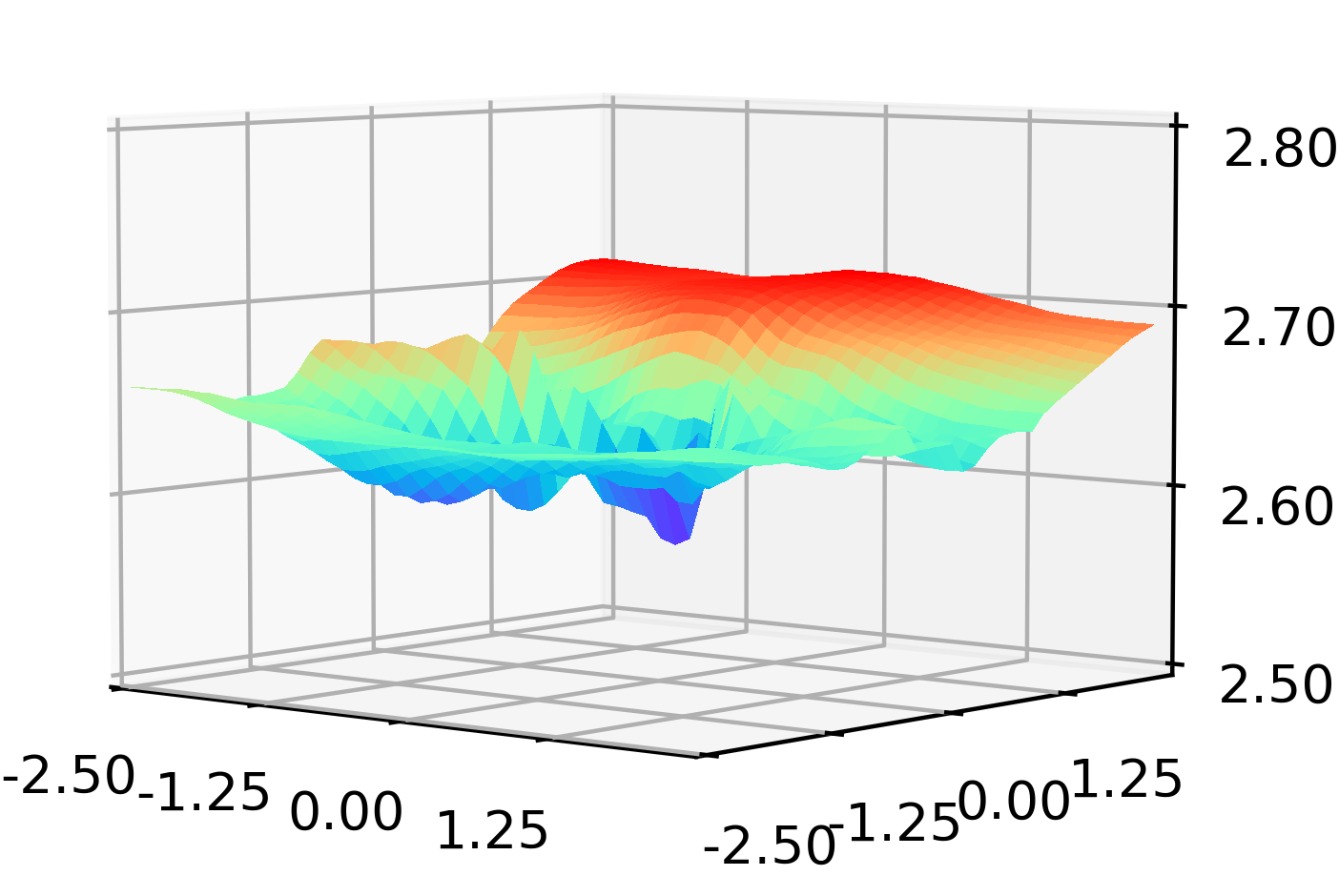}}  
\end{tabular}
& 
\begin{tabular}{@{\hskip 0.02in}c@{\hskip 0.02in} }
\parbox[c]{10em}{\includegraphics[width=10em]{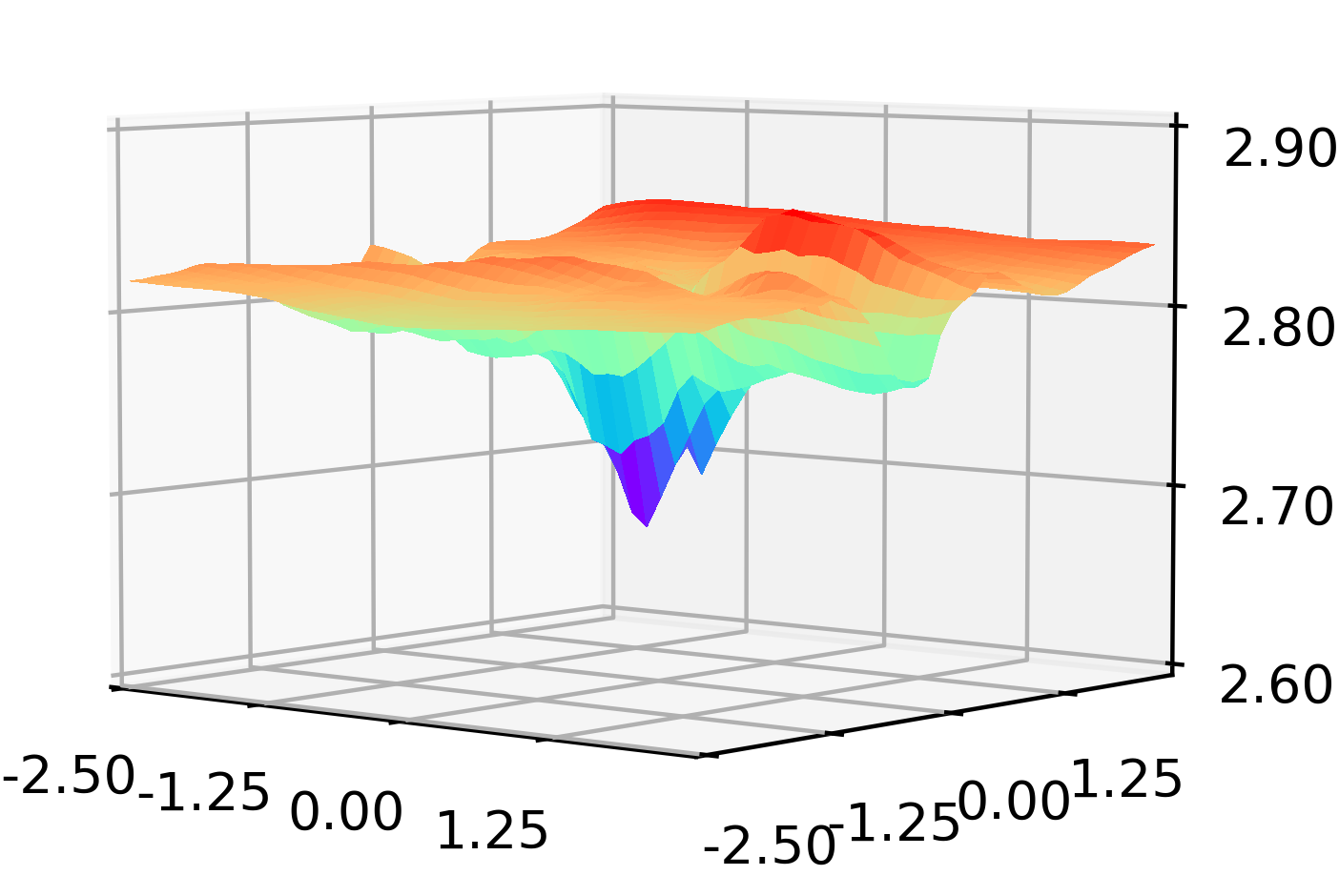}}  
\end{tabular}
\end{tabular}
\end{adjustbox}
\caption{\footnotesize{
Visualization of adversarial loss landscapes of {\FAT}, {\FATGA} and {\FBAT} trained using the ResNet-18 model on the CIFAR-10  dataset. 
The losses at are calculated w.r.t.  the same image example ID \#001456, and the landscape
is obtained by tracking the loss changes w.r.t. input variations following \cite{engstrom2018evaluating}.  
That is, the loss  landscape is generated by $z = \text{loss}(I + x \cdot \mathbf{r_1} + y \cdot \mathbf{r_2})$, where $I$ denotes an image, and the $x$-axis and the $y$-axis correspond to linear coefficients associated with the sign-based attack direction $\mathbf{r_1} = \text{sign}(
\nabla_{I} \text{loss}(I))$ 
$\mathbf{\hat{x}}$    
and  a random direction $\mathbf{r_2} \sim  \text{Rademacher}(0.5)$, respectively.
}}
\vspace{-3mm}
\label{fig: loss_landscape}
\end{figure*}

\begin{wraptable}{r}{55mm}
    \vspace*{-8mm}
    \caption{\footnotesize{SA and RA on ImageNet.
    }}
    \label{tab: imagenet}
    \vspace*{0.1in}
    \centering
    \vspace{1mm}
    \resizebox{0.3\textwidth}{!}{
    \begin{tabular}{c|c|c}
    \toprule[1pt]
    \midrule
         Method &  SA (\%) & RA-PGD (\%) \\
         \midrule
         {\FAT} & 60.90 
        %  \footnotesize{$\pm 0.47$} 
         & 43.43 
        %  \footnotesize{$\pm 0.24$} 
         \\
        %  {\ATF} &   60.89 & \textcolor{red}{\textbf{44.74}}\\
         {\FBAT} &  60.18
        %  \footnotesize{$\pm 0.09$} 
         & 44.64 
        %  \footnotesize{$\pm 0.11$} 
         \\
        %  \midrule
        %  {\FAT} & $4/255$ & 55.45 & 30.18 \\
        %  {\FBAT} & $4/255$ &  &  \\
    \midrule
    \bottomrule[1pt]
    \end{tabular}}
    \vspace{-3mm}
\end{wraptable}
We train DNN models ResNet (RN)-50~\citep{he2016deep} for ImageNet and we choose the training perturbation strength $\epsilon = 2/255$ strictly following \citep{Wong2020Fast,andriushchenko2020understanding}. We remark that when evaluating on ImageNet, we only compare ours with {\FAT} since as shown in Table~6 of \citep{andriushchenko2020understanding}, the other baseline methods did not show any improvement over Fast-AT at the attack budget $\epsilon = 2/255$. RA-PGD stands for the robustness against PGD-50-10 (50-step PGD attack with 10 restarts) with $\epsilon = 2/255$. Table\,\ref{tab: imagenet} shows the performance comparison between {\FAT} and {\FBAT}. We can see {\FBAT} outperforms {\FAT} by 1.23\% when facing attacks with $\epsilon = 2/255$. \RV{Since robust training on ImageNet usually takes small $\epsilon$ (like 2/255), the benefit of robust catastrophic overfitting alleviation becomes less evident.}

\paragraph{Sensitivity to regularization parameter $\lambda$
}

\begin{wraptable}{r}{75mm}
\vspace{-8mm}
\begin{center}
\caption{\footnotesize{Performance of {\FBAT} with different parameter $\lambda$. We train and evaluate with the same attack budget $\epsilon = 16/255$ on CIFAR-10 to show the influence brought by $\lambda$.} }
\label{tab: lambda_sensitivity}
\vspace*{0.1in}
\resizebox{0.4\textwidth}{!}{%
\begin{tabular}{cccccc}
\toprule[1pt]
\midrule
\multicolumn{6}{c}{CIFAR-10, PreActResNet-18, $\epsilon = 16/255$} \\ \midrule
\multicolumn{1}{c|}{1/$\lambda$~(/255)} & 500 & 1000 & 1500 & 2500 & 3500 \\ \midrule

\multicolumn{1}{c|}{SA~(\%)} & \textbf{83.20} & 75.06 & 69.31 & 68.16 & 64.37 \\

\multicolumn{1}{c|}{RA-PGD~(\%)} & 19.02 & 21.42 & 23.34 & \textbf{27.69} & 25.32\\ \midrule
 \bottomrule[1pt]
\end{tabular}%
}
\end{center}
\vspace{-3mm}
\end{wraptable}

In Table \ref{tab: lambda_sensitivity}, we show the sensitivity of {\FBAT} to the regularization parameter $\lambda$. All the parameters remain the same as the default setting, except that for different $\lambda$. We always adjust $\alpha_2$ so that $\alpha_2 / \lambda = 0.1\alpha_1$ holds. Note $1 / \lambda$ also serves as the attack step in  \eqref{eq: sol_linear}. As $\lambda$ decreases, the improvement in robust accuracy is evidently strengthened, and there is an obvious trade-off between robust accuracy (SA) and standard accuracy (RA). At a certain level of $\lambda$, namely when $\lambda \leq 255/3500$, RA starts to converge and stop surging. 

\paragraph{Sensitivity to different $\alpha_2$ choices}

\begin{wraptable}{r}{75mm}
\vspace{-8mm}
\centering
\caption{\footnotesize Performance of \FBAT with different $\alpha_2$ choices on CIFAR-10. Models are trained and evaluated with the same attack budget ($\epsilon=16/255$). Here $\alpha_1$ is set as the cyclic learning rate and is not a constant value. $\alpha_2$ is always set proportionate to $\alpha_1$ for simplicity.}
\label{tab: alpha_2_study}
\vspace*{0.1in}
\resizebox{.4\textwidth}{!}{%
\begin{tabular}{ccccc}
\toprule[1pt]
\midrule
\textbf{\begin{tabular}[c]{@{}c@{}}$\alpha_2$ (CIFAR-10, \\ PreActResNet18,\\ $\epsilon=16/255)$\end{tabular}} & 0.025$\alpha_1$ & 0.0167$\alpha_1$ & 0.0125$\alpha_1$ & 0.008$\alpha_1$ \\ \midrule
SA (\%) & \textbf{75.06} & 69.31 & 68.16 & 57.92 \\
RA-PGD (\%) & 21.42 & 23.34 & \textbf{27.69} & 20.53 \\ \midrule
\bottomrule[1pt]
\end{tabular}%
}
\vspace{-3mm}
\end{wraptable}

We consider the case of robust training with the large $\epsilon$ choice ($16/255$). As we can see from Table\,\ref{tab: alpha_2_study}, if $\alpha_2$ is set too small ($\alpha_2 = 0.008\alpha_1$), then both SA and RA will drop significantly. Here $\alpha_1$ is set as the cyclic learning rate and thus not a constant parameter. However, in the $\alpha_2$ interval $ [0.0125\alpha_1, 0.025\alpha_1]$, we observed a tradeoff between standard accuracy (SA) and robust accuracy (RA): That is, the improvement in RA corresponds to a loss in SA. In our experiments, we choose $\alpha_2$ when the tradeoff yields the best RA without suffering a significant drop of SA (which still outperforms the baseline approaches).

\paragraph{Sensitivity of linearization schemes}

Fast-BAT needs a good linearization point $\mathbf{z}$ in \eqref{eq: prob_biLevel_lin}. 
In experiments, we adopt the perturbation generated by 1-step PGD without sign as our default linearization scheme. In Table~\ref{tab: linearization_sensitivity}, we show the performance of the other possible linearization options. 
We find that 1-step PGD without sign achieves the best robust accuracy among all the choices. 
This is not spurring since this linearization point choice is consistent with the first-Taylor expansion that we used along the direction of the input gradient without the sign operation involved.
By contrast, {\FBAT} linearized with uniformly random noise suffers from catastrophic overfitting and reaches a rather low standard accuracy (SA).
{\FBAT} with other linearizations also yields a worse SA-RA trade-off than our proposal. 

\begin{wraptable}{r}{70mm}
\begin{center}
\vspace{-3mm}
\caption{\footnotesize{Performance of {\FBAT} with different linearization schemes. Besides 1-step PGD without sign (\textbf{PGD w/o Sign}), we further generate linearization point with the following methods: uniformly random noise $[-\epsilon, \epsilon]^d$~(\textbf{Uniformly Random}); uniformly random corner $\{-\epsilon, \epsilon \}^d$ }~(\textbf{Random Corner}); and perturbation from 1-step PGD attack with $0.5\epsilon$ as attack step~(\textbf{PGD}). 
}
\label{tab: linearization_sensitivity}
\vspace{0.1in}
\resizebox{0.4\textwidth}{!}{%
\begin{tabular}{cccccc}
\toprule[1pt]
\midrule
\multicolumn{5}{c}{CIFAR-10, PreActResNet-18, $\epsilon = 16/255$} \\ \midrule

\multicolumn{1}{c|}{\begin{tabular}[c]{@{}c@{}}Linearization \\ Method \end{tabular}} & \begin{tabular}[c]{@{}c@{}}PGD \\ w/o Sign \end{tabular} & \begin{tabular}[c]{@{}c@{}}Uniformly \\ Random \end{tabular} & \begin{tabular}[c]{@{}c@{}}Random \\ Corner \end{tabular} & PGD \\ \midrule
\multicolumn{1}{c|}{SA~(\%)} & 68.16 & 43.42 & 62.19 & \textbf{75.30} \\
\multicolumn{1}{c|}{RA-PGD~(\%)} & \textbf{27.69} & 21.25 & 16.5 & 19.42 \\ \midrule
 \bottomrule[1pt]
\end{tabular}%
}
\end{center}
\vspace{-5mm}
\end{wraptable}

\paragraph{The validity of the Hessian-free assumption on non-ReLU based neural networks.}
The Hessian-free assumption is based on the fact that the commonly used ReLU activation function is piece-wise linear \textit{w.r.t.} input. We further conduct experiments to verify the feasibility of such an assumption on models with non-ReLU activation functions. We choose two commonly used activation functions,  Swish \citep{ramachandran2017searching} and Softplus, as alternatives for the non-smooth ReLU function. We compare the results both calculating Hessian as well as the Hessian-free version to see if the Hessian-free assumption still holds for the non-ReLU neural network. The results are shown in Table~\ref{tab: activation_function}. As we can see, the use of Hessian does not affect performance much. A similar phenomenon can be observed across different $\epsilon$ and different model activation functions (ReLU, Softplus, and Swish). However, the introduction of Hessian leads to an increase in time consumption by more than 30\%. Therefore, we can draw the conclusion that the Hessian-free assumption is reasonable across different activation function choices. 

{\bf Ablation studies.} In Appendix~\ref{app: additional_result}, we   present additional empirical studies including 1) the sensitivity analysis of the linearization hyperparameter $\lambda$, 2) the choice of the linearization point, and 3) the sensitivity analysis of $\alpha_2$.

\paragraph{\RV{Comparisons with more baselines.}} 

In Tab.\,\ref{tab:baseline},
we compare {\FBAT} with more baselines, {\PGDS}, {\bs}, {\ATTA}, {\free}, and {\YOPO}. 
The standard PDG-7 AT typically yields the best RA, but causes the highest computation cost (see column `Time' for time till best model). While in the \textbf{fast} robust training paradigm, {\FBAT} stands top for different values of $\epsilon$.
{\FATGA} is indeed a strong baseline to {mitigate} robust catastrophic overfitting as train-time $\epsilon$ increases (e.g., $16/255$). Ours also outperforms {\free} and {\YOPO} in robust catastrophic overfitting alleviation as  $\epsilon$ grows. 
{\FATGA} paper also identified the incapability of {\free}.

\begin{table}[htb]
\centering
\caption{\footnotesize{Performance comparison  of {\FBAT} vs. baselines on (CIFAR10, PreActResNet-18). Each baseline  follows its original setting. 
{\free} (m=8) and \textsc{YOPO-5-3} are adopted.
\textsc{ATTA} refers to \textsc{ATTA-1-TRADES}.
For fair comparison, all the methods are trained with 20 epochs and cyclic learning rate. {Minor performance degradation compared to the results reported in the original papers is due to different training setting}, that we train all the methods with only 90\% of training data, choose the best model on the validation set (10\% training data), and evaluate on the test set. All the results are at the same level as the ones reported in the original papers (for those adopting similar training settings). 
Evaluation settings are consistent with  Table\,\ref{table: motivation_intro_overview} in the main paper.
}}
\resizebox{0.8\linewidth}{!}{
\begin{tabular}{c|ccc|ccc|c}
\toprule[1pt]
\multirow{2}{*}{Method} &
  \multicolumn{3}{c|}{$\epsilon=8/255$ (Train/test-time attack)} &
  \multicolumn{3}{c|}{$\epsilon=16/255$ (Train/test-time attack)} &
  \multirow{2}{*}{\begin{tabular}[c]{@{}c@{}}Time\\ (min)\end{tabular}} \\
      & \multicolumn{1}{c|}{SA}    & \multicolumn{1}{c|}{RA-PGD} & RA-AA & \multicolumn{1}{c|}{SA}    & \multicolumn{1}{c|}{RA-PGD} & RA-AA &      \\ \midrule
\textsc{AT} (\textsc{PGD-7}) &
  \multicolumn{1}{c|}{81.43\footnotesize{$\pm 0.13$}} &
  \multicolumn{1}{c|}{50.63\footnotesize{$\pm 0.16$}} &
  47.05\footnotesize{$\pm 0.18$} &
  \multicolumn{1}{c|}{61.55\footnotesize{$\pm 0.17$}} &
  \multicolumn{1}{c|}{31.11\footnotesize{$\pm 0.61$}} &
  22.99\footnotesize{$\pm 0.31$} &
  121.5 \\  \midrule
      
\FAT &
  \multicolumn{1}{c|}{{82.39}\footnotesize{$\pm 0.14$}} &
  \multicolumn{1}{c|}{45.49\footnotesize{$\pm 0.21$}} &
  41.87\footnotesize{$\pm 0.15$} &
  \multicolumn{1}{c|}{44.15\footnotesize{$\pm 7.27$}} &
  \multicolumn{1}{c|}{21.83\footnotesize{$\pm 1.32$}} &
  12.49\footnotesize{$\pm 0.33$} &
  \textbf{7.7} \\
\bs &
  \multicolumn{1}{c|}{79.31\footnotesize{$\pm 0.17$}} &
  \multicolumn{1}{c|}{48.06\footnotesize{$\pm 0.07$}} &
  44.55\footnotesize{$\pm 0.07$} &
  \multicolumn{1}{c|}{64.88\footnotesize{$\pm 1.75$}} &
  \multicolumn{1}{c|}{24.18\footnotesize{$\pm 1.37$}} &
  15.47\footnotesize{$\pm 0.92$} &
  20.2 \\
{\free} &
  \multicolumn{1}{c|}{79.59\footnotesize{$\pm 0.14$}} &
  \multicolumn{1}{c|}{42.84\footnotesize{$\pm 0.86$}} &
  39.39\footnotesize{$\pm 0.20$} &
  \multicolumn{1}{c|}{35.00\footnotesize{$\pm 12.37$}} &
  \multicolumn{1}{c|}{6.07\footnotesize{$\pm 1.95$}} &
  0.91\footnotesize{$\pm 0.42$} &
  24.5 \\
\FATGA &
  \multicolumn{1}{c|}{79.71\footnotesize{$\pm 0.24$}} &
  \multicolumn{1}{c|}{47.27\footnotesize{$\pm 0.22$}} &
  43.24\footnotesize{$\pm 0.27$} &
  \multicolumn{1}{c|}{58.29\footnotesize{$\pm 1.32$}} &
  \multicolumn{1}{c|}{26.01\footnotesize{$\pm 0.16$}} &
  17.97\footnotesize{$\pm 0.33$} &
  25.1 \\
\textsc{ATTA}
&
  \multicolumn{1}{c|}{79.43\footnotesize{$\pm 0.09$}} &
  \multicolumn{1}{c|}{48.78\footnotesize{$\pm 0.62$}} &
  44.61\footnotesize{$\pm 0.35$} &
  \multicolumn{1}{c|}{67.37\footnotesize{$\pm 1.89$}} &
  \multicolumn{1}{c|}{0.36\footnotesize{$\pm 0.12$}} &
  0.00\footnotesize{$\pm 0.00$} &
  38.8 \\
\textsc{YOPO-5-3} &
  \multicolumn{1}{c|}{\textbf{83.17}\footnotesize{$\pm 0.11$}} &
  \multicolumn{1}{c|}{44.50\footnotesize{$\pm 0.28$}} &
  40.61\footnotesize{$\pm 0.43$} &
  \multicolumn{1}{c|}{44.04\footnotesize{$\pm 3.61$}} &
  \multicolumn{1}{c|}{23.08\footnotesize{$\pm 2.30$}} &
  10.61\footnotesize{$\pm 0.86$} &
  43.8 \\
\rowcolor{Gray} \FBAT &
  \multicolumn{1}{c|}{79.97\footnotesize{$\pm 0.12$}} &
  \multicolumn{1}{c|}{\textbf{48.83}\footnotesize{$\pm 0.17$}} &
  \textbf{45.19}\footnotesize{$\pm 0.12$} &
  \multicolumn{1}{c|}{\textbf{68.16}\footnotesize{$\pm 0.25$}} &
  \multicolumn{1}{c|}{\textbf{27.69}\footnotesize{$\pm 0.16$}} &
  \textbf{18.79}\footnotesize{$\pm 0.24$} &
  20.5 \\ \bottomrule[1pt]
\end{tabular}
\label{tab:baseline}
}
\end{table}

\begin{table*}[htb]
\centering
\caption{\footnotesize Performance of {\FAT} and {\FBAT} with different activation functions on CIFAR-10. ReLU, Swish and Softplus are taken into consideration. For \FBAT, we compare the Hessian-free and Hessian-aware version to verify the influence of Hessian matrix. The results are averaged over 3 independent trials.}
\label{tab: activation_function}
\vspace*{0.1in}
\resizebox{.6\textwidth}{!}{%
\begin{tabular}{cccccc}
\toprule[1pt]
\midrule
\textbf{Setting} & \textbf{\begin{tabular}[c]{@{}c@{}}SA (\%)\\ ($\epsilon=8/255$)\end{tabular}} & \textbf{\begin{tabular}[c]{@{}c@{}}RA-PGD (\%)\\ ($\epsilon=8/255$)\end{tabular}} & \textbf{\begin{tabular}[c]{@{}c@{}}SA (\%)\\ ($\epsilon=16/255$)\end{tabular}} & \textbf{\begin{tabular}[c]{@{}c@{}}RA-PGD (\%)\\ ($\epsilon=16/255$)\end{tabular}} & \textbf{\begin{tabular}[c]{@{}c@{}}Time\\ (s/epoch)\end{tabular}} \\ \midrule

\footnotesize{Fast-AT-ReLU}
& 82.39\footnotesize{$\pm 0.14$} 
& 45.49\footnotesize{$\pm 0.21$} 
& 44.15\footnotesize{$\pm 7.27$} 
& 21.83\footnotesize{$\pm 1.32$} 
& 23.1 \\

\begin{tabular}[c]{@{}c@{}}\footnotesize{Fast-BAT-ReLU}\\\footnotesize{Hessian-free}\end{tabular} 
& 79.97 \footnotesize{$\pm 0.12$}
& 48.83 \footnotesize{$\pm 0.17$}
& 68.16 \footnotesize{$\pm 0.25$}
& 27.69 \footnotesize{$\pm 0.16$}
& 61.4 \\ 

\begin{tabular}[c]{@{}c@{}}\footnotesize{Fast-BAT-ReLU}\\\footnotesize{Hessian-aware}\end{tabular} 
& 79.62 \footnotesize{$\pm 0.17$}
& 49.13 \footnotesize{$\pm 0.14$}
& 67.82 \footnotesize{$\pm 0.23$}
& 27.82 \footnotesize{$\pm 0.19$}
& 82.6 \\ \midrule

\footnotesize{Fast-AT-Softplus} 
& 81.29 \footnotesize{$\pm 0.16$}
& 47.26 \footnotesize{$\pm 0.24$}
& 45.39 \footnotesize{$\pm 3.27$}
& 22.40 \footnotesize{$\pm 0.75$}
& 23.3 \\

\begin{tabular}[c]{@{}c@{}}\footnotesize{Fast-BAT-Softplus}\\ \footnotesize{Hessian-free}\end{tabular} 
& 79.48 \footnotesize{$\pm 0.18$}
& 49.67 \footnotesize{$\pm 0.21$}
& 68.57 \footnotesize{$\pm 0.27$}
& 25.59 \footnotesize{$\pm 0.15$}
& 61.7 \\ 

\begin{tabular}[c]{@{}c@{}}\footnotesize{Fast-BAT-Softplus}\\ \footnotesize{Hessian-aware}\end{tabular} 
& 79.59 \footnotesize{$\pm 0.21$}
& 49.74 \footnotesize{$\pm 0.12$}
& 68.63 \footnotesize{$\pm 0.23$}
& 25.54 \footnotesize{$\pm 0.19$}
& 82.8 \\ \midrule

\footnotesize{Fast-AT-Swish} 
& 75.61 \footnotesize{$\pm 0.15$}
& 44.43 \footnotesize{$\pm 0.18$}
& 52.03 \footnotesize{$\pm 4.29$}
& 23.08 \footnotesize{$\pm 2.23$}
& 23.1 \\

\begin{tabular}[c]{@{}c@{}}\footnotesize{Fast-BAT-Swish}\\\footnotesize{Hessian-free}\end{tabular} 
& 73.89 \footnotesize{$\pm 0.14$}
& 45.90 \footnotesize{$\pm 0.23$}
& 62.59 \footnotesize{$\pm 0.29$}
& 23.81 \footnotesize{$\pm 0.17$}
& 61.7 \\ 

\begin{tabular}[c]{@{}c@{}}\footnotesize{Fast-BAT-Swish}\\\footnotesize{Hessian-aware}\end{tabular} 
& 73.93 \footnotesize{$\pm 0.16$}
& 45.97 \footnotesize{$\pm 0.19$}
& 62.49 \footnotesize{$\pm 0.27$}
& 23.99 \footnotesize{$\pm 0.17$}
& 82.6 \\ \midrule
\bottomrule[1pt]
\end{tabular}%
}
\vspace{-5mm}
\end{table*}

\newpage
\section{Convergence Analysis}\label{sec: convergence}
\newcommand{\bx}{\mathbf{x}}
\newcommand{\by}{\mathbf{y}}
Let us consider we have data samples $\{\bx_i, \by_i \}_{i=1}^N$, where $N$ denotes the number of the training data. Then the goal of {\FBAT} algorithm is to solve:

{\small
\begin{equation}
    \begin{split}
    & \min_{\btheta} L_{\text{tr}}(\btheta, \bdelta^*(\btheta)) = \frac{1}{N} \sum_{i=1}^N \ell_{\text{tr}}(\btheta; \bx_i + \bdelta_i^*(\btheta), y_i) \\
    \bdelta^*(\btheta) \in & \argmin_{\bdelta_i \in \mathcal C} L_{\text{atk}} (\btheta, \bdelta^*(\btheta)) = \frac{1}{N} \sum_{i=1}^N \ell_{\text{atk}}(\btheta; \bx_i + \bdelta_i^*(\btheta), y_i)
\end{split}
\end{equation}}%

Let us denote the batch size as $b$. Then the problem can be reformulated as :

{\small
\begin{equation}
    \begin{split}
        & \min_{\btheta} L^{b_t}_{\text{tr}}(\btheta, \bdelta^*(\btheta)) = \frac{1}{b_t} \sum_{i=1}^{b_i} \ell_{\text{tr}}(\btheta; \bx_i + \bdelta_i^*(\btheta), y_i) \\
    \bdelta^*(\btheta) \in & \argmin_{\bdelta_i \in \mathcal C} L_{\text{atk}} (\btheta, \bdelta^*(\btheta)) = \underbrace{\frac{1}{b_t} \sum_{i=1}^{b_t} \ell_{\text{atk}}(\btheta; \bx_i + \bdelta_i^*(\btheta), y_i)}_{L_{\mathrm{atk}}^{b_t}(\btheta, \bdelta_i(\btheta))}
    \end{split}
    \label{eq: prob_biLevel_batch}
\end{equation}
}%
The way to computes the gradient \eqref{eq: GD_upper} is equal to the case where $b = N$.

\begin{align}
   \frac{d L^N_{\mathrm{tr}}(\boldsymbol \theta, \boldsymbol \delta^*(\boldsymbol \theta))}{d \btheta }
    = 
    \nabla_{\btheta} L^N_{\mathrm{tr}}(\boldsymbol \theta, \boldsymbol \delta^*(\btheta))
    + \frac{{d} \boldsymbol \delta^*(\boldsymbol \theta)^\top}{ {d} \boldsymbol \theta}
    \nabla_{\bdelta} L^N_{\mathrm{tr}}(\btheta, \bdelta^*(\btheta))
\end{align}

If we approximate the gradient $\nabla_{\btheta} L_{\text{tr}(\btheta, \bdelta^*(\btheta))}$ using a batch size of $b$, then we have the following assumptions:

\begin{itemize}
    \item Bias assumption:
    {\small
    \begin{align}
        \mathbb E [\nabla_{\btheta} L^b_{\text{tr}}(\btheta, \bdelta^*(\btheta))] = \nabla_{\btheta} L_{\text{tr}}(\btheta, \bdelta^*(\btheta)) + \beta(b),
    \end{align}}
    where $\beta(b)$ is the bias and the expectation is taken \textit{w.r.t.} batches. Note that $\beta(b) = 0$ for $b = N$.
    
    \item Variance assumption:
    {\small
    \begin{align}
        \mathbb E \| \nabla_{\btheta} L^b_{\text{tr}}(\btheta, \bdelta^*(\btheta)) - \nabla_{\btheta} L_{\text{tr}}(\btheta, \bdelta^*(\btheta)) - \beta(b) \|^2 = \sigma^2[1 + \| \nabla_{\btheta} L_{\text{tr}}(\btheta, \bdelta^*(\btheta)) \|^2],
    \end{align}}
    Note that the variance equals 0 for $b = N$.
\end{itemize}

Now the {\FBAT} algorithm with batch size $b$ can be stated as follows:

\begin{itemize}
    \item Initialization: $\btheta_0 \in \mathbb R^n$, step size $\{\alpha_t\}_{t=0}^{T-1}$, batch size $\{\alpha_b\}_{t=0}^{T-1}$.
    \item for $t = 1$ to $T$:
    \begin{itemize}
        \item Choose batsh size $b_t$ at each iteration $t \in [T]$.
        \item $L_{\text{tr}}(\btheta, \bdelta^*(\btheta)) \approx L^{b_t}_{\text{tr}}(\btheta, \bdelta^*(\btheta))$, $L_{\text{atk}}(\btheta, \bdelta^*(\btheta)) \approx L^{b_t}_{\text{atk}}(\btheta, \bdelta^*(\btheta))$
        \item Update:
        {\small
        \begin{align}
            \btheta_t = \btheta_{t-1} - \alpha_{t - 1}\nabla_{\btheta}L^{b_t}_{\text{tr}}(\btheta_{t-1}, \bdelta_{t-1}^*(\btheta_{t-1}))
        \end{align}}
        where
        {\small
        \begin{align}
            \nabla_{\btheta} L^{b_t}_{\text{tr}}(\btheta_{t-1}, \bdelta_{t-1}^*(\btheta_{t-1})) = \nabla_{\btheta} L^{b_t}_{\mathrm{tr}}(\boldsymbol \theta_{t-1}, \boldsymbol \delta^*(\btheta_{t-1}))
    + \nabla_{\btheta}\bdelta^*(\btheta_{t-1}) ^ \top
    \nabla_{\bdelta} L^{b_t}_{\mathrm{tr}}(\btheta_{t-1}, \bdelta^*(\btheta_{t-1})).
        \end{align}}
    Note $\nabla_{\btheta}\bdelta^*(\btheta_{t-1})$ is computed using \eqref{eq: IG_cons} by replacing:
    \begin{itemize}
        \item $L_{\text{atk}}(\btheta, \bdelta^*(\btheta))$ with $ L^{b_t}_{\text{atk}}(\btheta, \bdelta^*(\btheta))$
        \item $B_0$ with $B_0$ computed with \eqref{eq: prob_biLevel_batch}.
    \end{itemize}
    \end{itemize}
\end{itemize}

Assuming smoothness of $L_{\mathrm{tr}(\btheta)}$ \textit{w.r.t.} $\btheta$, we get:

{\small
\begin{align}
    L_{\mathrm{tr}}(\btheta_{t+1}) & \leq L_{\mathrm{tr}}(\btheta_{t}) + \left \langle \nabla_{\btheta}(\btheta_{t}), \btheta_{t+1} - \btheta_t \right \rangle + \frac{L}{2}\|\btheta_{t+1} - \btheta_{t}\|_2^2 \nonumber\\ 
    & = L_{\mathrm{tr}}(\btheta_{t}) - \alpha_t \left \langle \nabla_{\btheta}L_{\mathrm{tr}}(\btheta_{t}) , \nabla_{\btheta}L^{b_t}_{\mathrm{tr}}(\btheta_{t}) \right \rangle + \frac{\alpha_t^2 L}{2} \|\nabla_{\btheta}L^{b_t}_{\mathrm{tr}}(\btheta_{t})\|^2_2
\end{align}
}
Taking expectation \textit{w.r.t.} random samples, we get:

\newcommand{\bbe}{\mathbb{E}}
{\small
\begin{align*}
    \bbe [L_{\mathrm{tr}}(\btheta_{t+1})] & \leq \bbe[L_{\mathrm{tr}}(\btheta_{t})] 
    - \alpha_t \bbe [ \|\nabla_{\btheta} L_{\mathrm{tr}}(\btheta_{t})\|_2^2] 
    - \alpha_t \bbe [\left \langle \nabla_{\btheta}(\btheta_t), \beta(b_t) \right \rangle ] 
    + \alpha_t^2 \sigma^2 L \bbe[1 + \|\nabla_{\btheta} L_{\mathrm{tr}}(\btheta_t)\|^2_2] \\
    & \hspace{6cm} + 2\alpha_t^2 L \bbe[ \|\nabla_{\btheta} L_{\mathrm{tr}}(\btheta_t)\|^2_2] 
    + 2\alpha_t^2 L\bbe[\|\beta(b_t)\|^2_2] \\
    % \bbe\left[ L_{\mathrm{tr}}(\btheta_t) - \alpha_t \| \nabla_{\btheta}L_{\mathrm{tr}}(\btheta_t) - \frac{\alpha_2}{2}\|\nabla_{\btheta}L_{\mathrm{tr}}(\btheta_t) + \beta(b_t)\|^2 + \frac{\alpha_t}{2} \|_2^2 \right]
    \\
    & \leq 
    \bbe\left[L_{\mathrm{tr}}(\btheta_{t}) 
    - \left[ \frac{\alpha_t}{2} - \alpha_t^2 \sigma^2L \right] \|\nabla_{\btheta} L_{\mathrm{tr}}(\btheta_t)\|_2^2 + \left[ \frac{\alpha_t}{2} + 2\alpha_t^2L \right]\|\beta(b_t)\|^2_2  + \alpha^2_t L \sigma^2 \right]
\end{align*}
}

Using $\alpha_t \leq \frac{1}{4L(2 + \sigma^2)}$, we get:

{\small
\begin{align}
    \bbe[L_{\mathrm{tr}}(\btheta_{t+1})] \leq \bbe\left[L_{\mathrm{tr}}(\btheta_{t}) - \frac{\alpha_t}{4} \|\nabla_{\btheta} L_{\mathrm{tr}}(\btheta_{t})\|^2_2 + \alpha_t \|\beta(b_t)\|^2 + \alpha_t^2 L \sigma^2 \right].
\end{align}
}

Finally, after rearranging the terms, we get:

{\small
\begin{align}
    \frac{\alpha_t}{4} \bbe[\|\nabla_{\btheta}L_{\mathrm{tr}}(\btheta_{t})\|_2^2] 
    \leq
    \bbe[(L_{\mathrm{tr}}(\btheta_{t}) - L_{\mathrm{tr}}(\btheta_{t+1})) + \alpha_t \|\beta(b_t)\|^2 + \alpha_t^2 L \sigma^2 ]
\end{align}
}

Summing over $t = 1$ to $T$ and multiplying by $1/T$, we get:

{\small
\begin{align}
    \frac{1}{T} \sum_{t=1}^T \frac{\alpha_t}{4}\bbe[\|\nabla_{\btheta}L_{\mathrm{tr}}(\btheta_{t})\|^2_2] \leq \frac{L_{\mathrm{tr}}(\btheta_{0}) - L_{\mathrm{tr}}(\btheta^*)}{T} + \frac{1}{T}\sum_{t=1}^T \alpha_t \|\beta(b_t)\|^2 + \frac{1}{T}\sum_{t=1}^T \alpha_t^2 \sigma_t^2.
\end{align}
}

Taking $\alpha_t = \alpha$ and $b_t = b$, we get:

{\small
\begin{align}
    \frac{1}{T}\sum_{t=1}^T\bbe[\|\nabla_{\btheta}L_{\mathrm{tr}}(\btheta_{t})\|_2^2] \leq \frac{4[L_{\mathrm{tr}}(\btheta_{0}) - L_{\mathrm{tr}}(\btheta^*)]}{\alpha T} + 4 \|\beta(b)\|^2 + 4\alpha \sigma^2.
\end{align}
}

Recall if $b=N$, we have $\sigma^2 = 0$ and $\beta(b) = 0$, we can further get:

{\small
\begin{align}
    \frac{1}{T}\sum_{t=1}^T\|\nabla_{\btheta}L_{\mathrm{tr}}(\btheta_{t})\|_2^2 \leq \frac{4[L_{\mathrm{tr}}(\btheta_{0}) - L_{\mathrm{tr}}(\btheta^*)]}{\alpha T}.
\end{align}}

Case I: We can choose $\alpha = \frac{1}{8L}$ and get:

{\small
\begin{align}
    \frac{1}{T}\|\nabla_{\btheta}L_{\mathrm{tr}}(\btheta_{t})\|_2^2 = \mathcal O (\frac{1}{T}).
\end{align}}

Case II: We can choose $\alpha = \sqrt{\frac{1}{T}}$:

{\small
\begin{align*}
    \frac{1}{T} \sum_{t=1}^T \bbe[\|\nabla_{\btheta}L_{\mathrm{tr}}(\btheta_{t}\|^2_2)] \leq \frac{4[L_{\mathrm{tr}}(\btheta_{0}) - L_{\mathrm{tr}}(\btheta^*)]}{\sqrt{T}} + \underbrace{\frac{4\sigma^2}{\sqrt{T}}}_{\mathcal{O}(\frac{1}{\sqrt{T}})} + \underbrace{4\|\beta(b)\|^2}_{\mathcal{O}(1)}
\end{align*}}

Now if $\|\beta(b)\|^2 = \mathcal O(\frac{1}{\sqrt{T}})$, then

{\small
\begin{align}
    \frac{1}{T}\sum_{t=1}^T \bbe[\|\nabla_{\btheta}L_{\mathrm{tr}}(\btheta_{t}\|^2_2] \leq \mathcal{O}(\frac{1}{\sqrt{T}}),
\end{align}}

otherwise we get in the worst case:

{\small
\begin{align}
    \frac{1}{T}\sum_{t=1}^T \bbe[\|\nabla_{\btheta}L_{\mathrm{tr}}(\btheta_{t}\|^2_2] \leq \mathcal{O}(\frac{1}{\sqrt{T}}) + \mathcal{O}(1).
\end{align}}

\end{document}